\documentclass[10pt,twocolumn,letterpaper]{article}

\usepackage[pagenumbers]{cvpr} 
\usepackage{multirow} 
\usepackage{algorithm}
\usepackage{algpseudocode}

\usepackage{booktabs}      
\usepackage{graphicx}      
\usepackage[table]{xcolor} 
\colorlet{lightgray}{gray!20} 
\definecolor{cvprblue}{rgb}{0.21,0.49,0.74}
\usepackage[pagebackref,breaklinks,colorlinks,allcolors=cvprblue]{hyperref}

\def\paperID{39766} 
\def\confName{CVPR}
\def\confYear{2026}

\title{DeAR: Fine-Grained VLM Adaptation by Decomposing Attention Head Roles}

\author{
    Yiming Ma\textsuperscript{1,*} \quad
    Hongkun Yang\textsuperscript{2,*} \quad
    Lionel Z. Wang\textsuperscript{3,\dag} \quad
    Bin Chen\textsuperscript{1,4,\dag} \quad
    Weizhi Xian\textsuperscript{1} \quad
    Jianzhi Teng\textsuperscript{3} \\
    \textsuperscript{1}Chongqing Research Institute of Harbin Institute of Technology, China \\
    \textsuperscript{2}Ocean University of China, China \\
    \textsuperscript{3}The Hong Kong Polytechnic University, Hong Kong SAR \\
    \textsuperscript{4}Harbin Institute of Technology (Shenzhen), China \\
}

\begin{document}
\maketitle
\def\thefootnote{*}\footnotetext{Equal contribution. \quad \textsuperscript{\dag}Corresponding authors.}
\def\thefootnote{\arabic{footnote}} 

\begin{abstract}
Prompt learning is a dominant paradigm for adapting pre-trained Vision-Language Models (VLMs) to downstream tasks. However, existing methods often rely on a simplistic, layer-centric view, assuming shallow layers capture general features while deep layers handle task-specific knowledge. This assumption results in uncontrolled interactions between learnable tokens and original tokens. Task-specific knowledge could degrade the model's core generalization and creates a trade-off between task adaptation and the preservation of zero-shot generalization. To address this, we challenge the layer-centric view and propose \textbf{DeAR}, a framework that achieves fine-grained VLM adaptation by \textbf{De}composing \textbf{A}ttention head \textbf{R}oles. We posit that the functional specialization within VLMs occurs not between layers, but at the finer-grained level of individual attention heads in the deeper layers. Based on this insight, we introduce a novel metric, Concept Entropy, to systematically classify attention heads into distinct functional roles: \textit{Attribute}, \textit{Generalization}, and \textit{Mixed}. Guided by these roles, we introduce specialized attribute tokens and a Role-Based Attention Mask mechanism to precisely control information flow, ensuring generalization heads remain isolated from task-specific knowledge. We further incorporate a Task-Adaptive Fusion Strategy for inference. Extensive experiments on fifteen datasets show that DeAR achieves a strong balance between task adaptation and generalization, outperforming previous methods across various tasks. Our code is available at \url{https://github.com/wellsssssss/DeAR}
\end{abstract}
\section{Introduction}

Pre-trained Vision-Language Models (VLMs), such as CLIP \citep{radford2021learningtransferablevisualmodels}, have emerged as a new paradigm in computer vision, demonstrating remarkable zero-shot generalization capabilities across a wide range of downstream tasks. By leveraging contrastive learning on over 400 million image-text pairs, CLIP learns to align visual and textual inputs in a shared embedding space. This has led to impressive achievements in tasks such as text-image retrieval \citep{radford2021learningtransferablevisualmodels}, visual question answering (VQA) \citep{gao2025clipadapterbettervisionlanguagemodels}, and open-vocabulary detection and segmentation \citep{gu2022openvocabularyobjectdetectionvision,Wang_Chen_Kang_Li_Xian_Chen_Xu_2025}. Moreover, the impact of CLIP extends to traditional vision tasks, such as image classification and object detection \citep{9879567}, where its zero-shot performance can surpass that of fully supervised methods.

Despite these successes, adapting CLIP to specific downstream tasks remains a significant challenge. Its pre-trained knowledge often lacks the granularity required for specialized domains. Although full fine-tuning could incorporate this new knowledge, it is computationally expensive and often leads to "catastrophic forgetting," where the model loses its powerful zero-shot generalization abilities \citep{kirkpatrick2017overcoming}. Consequently, Parameter-Efficient Fine-Tuning (PEFT) methods have become the mainstream approach \citep{wang2024megafaketheorydrivendatasetfake, liu2025sara}. These methods keep the VLM backbone frozen and introduce a small number of trainable parameters. PEFT strategies include adapter-based methods and prompt learning. The former typically insert lightweight MLP layers to refine the output features \citep{seputis2024multimodaladaptervisionlanguagemodels,gao2025clipadapterbettervisionlanguagemodels,zhu2023featuresmatterenhancingfewshot}, while prompt learning has emerged as a dominant and effective paradigm.

\begin{figure}[t]
    \centering
    \includegraphics[width=0.5\textwidth]{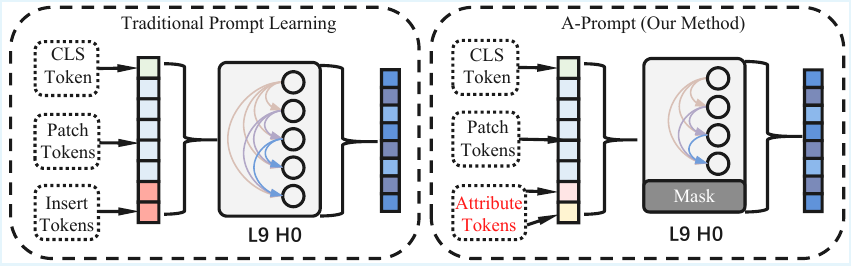}
    \caption{For a Generalization Head (e.g., Layer 9 Head 0), our Role-Based Mask explicitly blocking the original CLS and Patch tokens from interacting to the inserted Attribute Tokens.}
    \label{fig:compare}
\end{figure}

Prompt learning for VLMs was pioneered by CoOp \citep{Zhou_2022}, which introduced "soft prompts" , learnable continuous vectors, into the text encoder to adapt to downstream tasks. Subsequent work extended this concept to the vision domain, with VPT \citep{jia2022visualprompttuning} demonstrating the effectiveness of visual prompting. A key breakthrough came with MaPLe \citep{khattakMaPLe}, which proposed the first multi-modal prompt learning framework, tuning prompts in both the vision and text encoders and proving its superiority over uni-modal approaches.

However, even with a frozen backbone, prompt learning methods still risk losing generalization capabilities. Due to the nature of the Vision Transformer \citep{dosovitskiy2021imageworth16x16words,vaswani2023attentionneed}, inserted tokens inevitably interact with original tokens via multi-head self-attention, which can harm generalization. Existing solutions attempt to address this by carefully selecting which layers to inject prompts into—some targeting early layers (MaPLe), while others target deeper ones (MMRL). These conflicting, layer-based strategies reveal a fundamental gap: they treat entire Transformer layers as black boxes, ignoring the diverse functional roles played by individual attention heads within them. This lack of a fine-grained, principled approach for integrating new knowledge results in the challenging trade-off between achieving effective adaptation and maintaining generalization.


To address these challenges, we draw inspiration from recent work on VLM interpretability \citep{gandelsman2024interpreting} and propose a novel framework: \textbf{DeAR}, a novel framework for fine-grained VLM adaptation achieved by \textbf{De}composing \textbf{A}ttention head \textbf{R}oles. Given that CLIP's final image representation is constructed by its late attention layers, we suppose that the key to effective task adaptation, while preserving generalization, is to focus on the functional specialization of these individual heads.  Our approach begins by quantitatively analyzing the function of each attention head in the later layers of a ViT-B-16 backbone \citep{dosovitskiy2021imageworth16x16words}. Using our proposed Concept Entropy metric, we classify these heads into distinct functional roles: Attribute Heads, Generalization Heads, and Mixed Heads. Leveraging this functional classification, DeAR introduces specialized, learnable attribute tokens and employs a role-based Attention Mask strategy. This mechanism precisely controls the information flow, guiding attribute tokens to interact primarily with their corresponding expert heads, while critically isolating the generalization heads from any new task-specific knowledge (Figure~\ref{fig:compare}). Our contributions are as follows:
\begin{itemize}
    \item We introduce \textbf{Concept Entropy}, a novel quantitative metric to systematically analyze and classify the functional roles of attention heads in ViT-B/16, revealing a clear specialization into Attribute, Generalization, and Mixed roles.
    \item We propose the \textbf{DeAR}, which introduces a Role-Based Attention Mask to leverage this functional map. This mechanism achieves controllable fine-tuning by precisely routing new knowledge to expert heads while shielding generalization heads, achieving a balance between task-specific adaptation and the preservation of generalization capabilities.
    \item We demonstrate through extensive experiments that DeAR achieves \textbf{new state-of-the-art performance} on the challenging base-to-novel generalization benchmark.
\end{itemize}
\section{Related Work}

\subsection{Vision-Language Models (VLMs)}
Foundational Vision-Language Models (VLMs) have revolutionized multimodal learning. This paradigm was pioneered by CLIP \citep{radford2021learningtransferablevisualmodels}, which established a powerful contrastive learning framework on massive web-scale data. Subsequent efforts have expanded on this foundation in diverse ways: ALIGN \citep{pmlr-v139-jia21b} demonstrated the power of training on even larger, noisier datasets; Florence \citep{yuan2021florencenewfoundationmodel} aimed for a universal visual representation via multi-task pre-training; SigLIP \citep{10377550} enhanced scalability with a simpler pairwise sigmoid loss; and more recently, OpenVision \citep{li2025openvisionfullyopencosteffectivefamily} has proposed full transparency by leveraging high-quality, LLM-generated synthetic captions.

\subsection{Prompt Learning for VLMs}
Prompt learning has emerged as a dominant parameter-efficient strategy for adapting VLMs. The line of research was pioneered by CoOp, which replaces hand-crafted templates with learnable prompt vectors\citep{Zhou_2022}. However, CoOp learns a single static prompt for the entire dataset, which tends to overfit seen classes and lacks adaptability to individual images. Subsequent studies proposed CoCoOp, which conditions prompts on each image, and Task-Aware Clustering, which learns cluster-specific prompts for semantic subgroups\citep{zhou2022cocoop,Hao_2025_CVPR}. To preserve the general knowledge of the backbone, PromptSRC, ProGrad and TCP introduce various regularization techniques\citep{Khattak_2023_ICCV,https://doi.org/10.48550/arxiv.2205.14865,TCP24}.

As research matured, the focus expanded from uni-modal text prompts toward multimodal strategies. MaPLe presents a framework for simultaneous prompting in both vision and language branches\citep{khattakMaPLe}, and more recent work such as MMRL designs shared representation spaces to facilitate deeper cross-modal interactions\citep{guo2025mmrl}. The latest trend pursues fine-grained semantics: LLaMP employs LLMs as adaptive prompt learners, ATPrompt constructs hybrid attribute--category prompt structures, PromptKD exploits prompts for knowledge distillation, and TextRefiner mines internal visual knowledge of VLMs to refine prompts without external models\citep{Zheng_2024_Large,li2025advancing,li2024promptkd,xie2024textrefinerinternalvisualfeature}. Alternatively, Skip Tuning forgoes additional parameters altogether and instead modifies the full fine-tuning baseline with layer-wise and class-wise skipping to improve knowledge transfer \citep{wu2024skip}.

\begin{figure*}[t]
    \centering
    \includegraphics[width=0.82\textwidth]{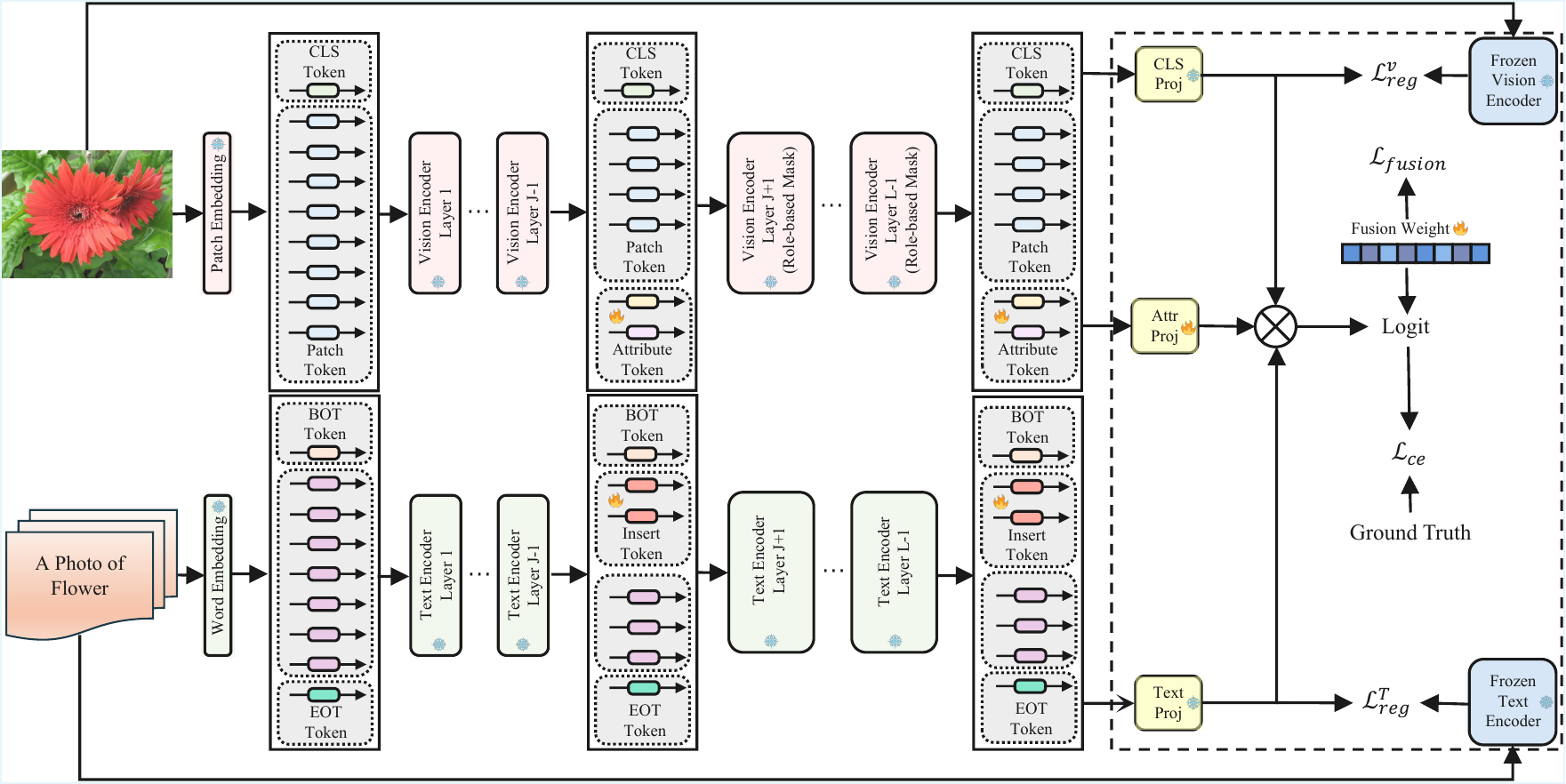}
    \caption{The overall framework of DeAR. At a deep layer $J$, we insert learnable attribute tokens into the frozen vision encoder and the text encoder. Information flow in these subsequent layers is precisely controlled by our Role-Based Attention Mask mechanism. For inference, the class feature (from the \texttt{[CLS]} token) and attribute feature first compute the logits with the text feature. These logits are then combined via learnable fusion weights to produce the final prediction.}
    \label{fig:framework}
\end{figure*}

\section{Method}
Despite advances in VLMs and prompt learning, efficiently adapting these models while maintaining generalization is still challenging. Most existing methods focus on designing representation spaces or using prompts to inject external knowledge, but rarely explore how internal information flows within the model. Our work, DeAR, addresses this by directly investigating these internal interactions to enable attribute-aware and interpretable prompt learning.
\subsection{Revisiting the Vision Transformer}
\label{sec:preliminaries}

The ViT-based image encoder first converts an image into a sequence of patch embeddings. An input image $\mathbf{x}$ is divided into $N$ patches, which are linearly projected into $D$-dimensional vectors. A learnable \texttt{[CLS]} token, $\mathbf{z}_{\text{cls}}$, is prepended to this sequence, and position embeddings $\mathbf{E}_{\text{pos}}$ are added to yield the initial token sequence $\mathbf{z}^0 \in \mathbb{R}^{(N+1) \times D}$.


This sequence is then processed by a stack of $L$ Transformer layers. Each layer consists of a Multi-Head Self-Attention (MSA) module and a feed-forward MLP, with residual connections and layer normalization. Central to our work, the MSA mechanism allows tokens to interact through $h$ parallel "attention heads." The MSA module's output is computed by concatenating the outputs of all heads, i.e., $\text{MSA}(\mathbf{z}) = \text{Concat}(\text{head}_1, \dots, \text{head}_h)\mathbf{W}_o$, where $\mathbf{W}_o$ is a final projection layer. Each individual head is computed via scaled dot-product attention:
\begin{equation}
\text{head}_i = \text{softmax}\left(\frac{\mathbf{Q}_i\mathbf{K}_i^T}{\sqrt{D_h}}\right) \mathbf{V}_i
\label{eq:single_head_attention}
\end{equation}

\begin{table*}[t]
\centering
\caption{Representative examples of functionally specialized Attribute Heads identified in the later layers of ViT-B-16. Each head demonstrates a strong and unambiguous focus on a single core visual attribute, as evidenced by its top descriptive phrases.}
\resizebox{\textwidth}{!}{%
\begin{tabular}{l|l|l|l|l}
\toprule
\textbf{Object Head} & \textbf{Shape Head} & \textbf{Texture Head} & \textbf{Color Head} & \textbf{Location Head} \\
\textbf{(L12, H7)} & \textbf{(L12, H5)} & \textbf{(L11, H5)} & \textbf{(L12, H10)} & \textbf{(L12, H0)} \\
\midrule
- A stick & - A regular octagon & - Delicate embroidery & - An image with cold green tones & - Picture taken in Scotland \\
- A cup & - Rectangular object & - Intricate architectural carving & - Image with a red color & - Photo taken in Barcelona \\
- A bonnet & - A wavy pattern & - Artwork with kaleidoscopic patterns & - A charcoal gray color & - Photo taken in Tokyo, Japan \\
- A trampoline & - Herringbone pattern & - Image with woven basket textures & - Image with a yellow color & - Image taken in South Africa \\
- A shoelace & - A scalene triangle & - Artwork with Mondrian-like grids & - Image with a blue color & - Picture taken in Texas, USA \\
- A bottle & - A polka dot & - Image with calligraphy writing & - A platinum silver color & - Photo taken in Paris, France \\
\bottomrule
\end{tabular}%
}
\label{tab:attribute_head_examples_wide}
\end{table*}

The Query ($\mathbf{Q}_i$), Key ($\mathbf{K}_i$), and Value ($\mathbf{V}_i$) matrices for each head $i$ are learned linear projections of the input tokens $\mathbf{z}$ using per-head weight matrices $(\mathbf{W}_{qi}, \mathbf{W}_{ki}, \mathbf{W}_{vi})$. 
After passing through all $L$ layers, the final state of the class token, $\mathbf{z}_{\text{cls}}^L$, is projected to yield the final image feature $\mathbf{f}_{\text{cls}}$. This multi-head self-attention mechanism forms the foundation of our controllable fine-tuning framework.

\subsection{Identifying Functional Roles of Attention Heads}
\label{sec:head_roles}

The core of DeAR is our hypothesis, inspired by recent interpretability studies \citep{gandelsman2024interpreting}, that the functional specialization of a VLM resides not at the layer level, but within individual attention heads in the model's deeper layers \citep{10.5555/3540261.3541188}. We propose a novel, unsupervised methodology to first \textit{identify} and then \textit{quantify} the functional roles of these heads \citep{vig-belinkov-2019-analyzing}. Our analysis focuses on the later layers (9--12) of ViT-B-16, which are known to be critical for constructing the final image representation.

To avoid subjective, pre-defined attribute categories, we adopt an approach to automatically identify conceptual clusters from the heads' description. Following \citet{gandelsman2024interpreting}, we first use TEXTSPAN to generate a ranked list of top-$N$ descriptive text phrases for each attention head. This yields a rich corpus of phrases describing each head's attentional focus. We then employ a pre-trained SBERT, $S(\cdot)$ \citep{reimers-gurevych-2019-sentence}, to map each phrase into a high-dimensional semantic space. Instead of classifying these phrase embeddings into a fixed set of categories, we perform density-based clustering using HDBSCAN \citep{mcinnes2017hdbscan}, a robust algorithm that can discover clusters of varying shapes and densities without requiring a pre-specified number of clusters. This automated process reveals a diverse set of conceptual clusters directly from the data. 

Our clustering analysis revealed 12 distinct conceptual groups, including concepts such as \textit{location}, \textit{text}, \textit{object}, \textit{animal}, \textit{photographic style}, and \textit{emotion}, etc. From these, we selected five frequent and semantically orthogonal concepts most relevant for general visual recognition tasks: $A = \{\texttt{color}, \texttt{shape}, \texttt{texture}, \texttt{object}, \texttt{location}\}$. This process provides the conceptual basis for our subsequent analysis, where the centroid of each selected cluster $c_j$ is denoted by its embedding $\bar{\mathbf{s}}_j = S(c_j)$.

\paragraph{Quantifying Specialization with Concept Entropy.}
With the core attributes established, we introduce a novel metric, \textbf{Concept Entropy}, to quantify the degree of functional specialization for each attention head $(l,h)$. For its top-$N$ descriptive phrases $D_{(l,h)}$, we first classify each phrase $d_i$ to the closest attribute cluster $c_j \in C$. This yields a probability distribution $P_{(l,h)}$ over the attributes, where $P_{(l,h)}(c_j)$ is the fraction of phrases assigned to $c_j$. Finally, the Concept Entropy $H(P_{(l,h)})$ is the Shannon entropy of this distribution. These steps are defined as:
\begin{gather}
    \text{cat}(d_i) = \arg\max_{c_j \in C} \cos\big(S(d_i), \bar{\mathbf{s}}_j\big) \\
    P_{(l,h)}(c_j) = \frac{1}{N} \sum\nolimits_{i=1}^{N} \mathbb{I}\big(\text{cat}(d_i) = c_j\big) \label{eq:prob_dist} \\ 
    H(P_{(l,h)}) = -\sum\nolimits_{j=1}^{k} P_{(l,h)}(c_j) \log_2 P_{(l,h)}(c_j)
\end{gather}
where $\mathbb{I}(\cdot)$ is the indicator function.
A low Concept Entropy score signifies that a head is highly specialized, concentrating its function on a single core attribute (\textbf{Attribute Head}). Conversely, a high score indicates a generalist function, engaging with a wide range of abstract concepts (\textbf{Generalization Head}). Heads with intermediate scores are classified as \textbf{Mixed Heads}. This role identification framework forms the foundation for DeAR's controllable learning mechanism.

\subsection{DeAR: Fine-Grained Attribute-Aware Prompt Learning}
\label{sec:a_prompt_framework}

Building on our function analysis of attention heads, we introduce DeAR, a framework consisting of three key components: multimodal attribute-aware prompt learning, a role-based attention mask mechanism, and a task-adaptive inference strategy. The overall framework is shown in Figure~\ref{fig:framework}.

\subsubsection{Multimodal Attribute-Aware Prompting}
To adapt CLIP for downstream tasks, DeAR introduces learnable tokens into both the vision and text encoders, ensuring new knowledge is aligned across modalities.

\paragraph{Vision-Side Prompting.}
We initialize a set of learnable "attribute tokens," $\{\mathbf{r}_{\text{attr}} \in \mathbb{R}^{D_v}\}_{\text{attr} \in A}$, where $D_v$ is the feature dimension of the vision encoder. Here, $A$ is the set of core visual attributes identified in our analysis, specifically $A = \{\text{color}, \text{shape}, \text{texture}, \text{object}, \text{location}\}$. Based on the findings of \citet{gandelsman2024interpreting} that the final image representation in CLIP is predominantly constructed by the later attention layers, these attribute tokens are integrated into the higher layers of the image encoder $\mathcal{V}$, beginning from the $J$-th layer (we set $J$ = 9 in our experiments, following prior work on ViT analysis).

For the image encoder $\mathcal{V}$, the forward pass through the first $J-1$ layers proceeds as standard:
\begin{align*}
    [\mathbf{z}_{\text{cls}}^{\ell}, \mathbf{z}_{p}^{\ell}] &= \mathcal{V}_{\ell}([\mathbf{z}_{\text{cls}}^{\ell-1}, \mathbf{z}_{p}^{\ell-1}]) \quad \ell = 1, \dots, J-1
\end{align*}
For subsequent layers $\ell \ge J$, the process involves the attribute tokens. Let $\mathbf{r}_{\text{attr}}^{\ell-1}$ be the state of an attribute token before entering layer $\ell$. The Transformer layer $\mathcal{V}_{\ell}$ processes the augmented sequence to produce an updated, context-aware state $\tilde{\mathbf{r}}_{\text{attr}}^{\ell}$. The input to the \textit{next} layer, $\mathbf{r}_{\text{attr}}^{\ell}$, is then determined by a controlled mixture:
\begin{align}
&[\mathbf{z}_{\text{cls}}^{\ell}, \mathbf{z}_{p}^{\ell}, \{\tilde{\mathbf{r}}_{\text{attr}}^{\ell}\}_{\text{attr} \in A}] 
= \mathcal{V}_{\ell}([\mathbf{z}_{\text{cls}}^{\ell-1}, \mathbf{z}_{p}^{\ell-1}, \{\mathbf{r}_{\text{attr}}^{\ell-1}\}_{\text{attr} \in A}]) \nonumber \\
&[\mathbf{z}_{\text{cls}}^{\ell+1}, \mathbf{z}_{p}^{\ell+1}, \{\tilde{\mathbf{r}}_{\text{attr}}^{\ell+1}\}_{\text{attr} \in A}] \nonumber \\
&\quad = \mathcal{V}_{\ell+1}([\mathbf{z}_{\text{cls}}^{\ell}, \mathbf{z}_{p}^{\ell}, \{\beta \mathbf{r}_{\text{attr}}^{\ell}+ (1 - \beta) \tilde{\mathbf{r}}_{\text{attr}}^{\ell}\}_{\text{attr} \in A}])
\label{eq:knowledge_anchoring_vision}
\end{align}
Inspired by MMRL++ \citep{guo2025mmrlparameterefficientinteractionawarerepresentation}, we control the information flow between layers using the hyperparameter $\beta \in [0, 1]$. A purely sequential evolution ($\beta=0$) allows tokens to adapt fully to the context of vision feature, but risks "semantic drift," where the token deviates from its core learned meaning. Conversely, a static re-injection of the raw token at each layer ($\beta=1$) strongly preserves the core meaning but prevents any contextual adaptation. Our approach Eq. (\ref{eq:knowledge_anchoring_vision}) strikes a flexible balance: it allows the attribute token to absorb image-specific information via $\tilde{\mathbf{r}}_{\text{attr}}^{\ell}$ while being regularized by its foundational prior $\mathbf{r}_{\text{attr}}$. This prevents the representation from deviating excessively from its core attribute meaning, leading to more stable prompt learning.

\paragraph{Text-Side Prompting.}
To ensure cross-modal alignment, we apply a symmetrical strategy to the text encoder $\mathcal{W}$. We initialize a set of $K$ learnable tokens, $\{\mathbf{p}_k\}$, which are injected from layer $J$ to $L$ between the [BOT] and word tokens. Since text prompts for classification are typically simple templates (e.g., ``a photo of \{cls\}''), a small set of parameters suffices for alignment without needing attribute-specific designs. The forward pass through the text encoder is then modified. For any given layer $\ell$, let the full sequence of input tokens be $\mathbf{T}^{\ell-1}$. The output is computed as $\mathbf{T}^{\ell} = \mathcal{W}_{\ell}(\mathbf{T}^{\ell-1})$, where the prompt tokens within $\mathbf{T}^{\ell-1}$ are defined conditionally:
\begin{equation}
\label{eq:knowledge_anchoring_text}
\mathbf{p}_k^{\text{in}, \ell} = 
\begin{cases} 
    \mathbf{p}_k^{\ell-1} & \text{if } \ell < J \\
    \beta \mathbf{p}_k + (1-\beta) \tilde{\mathbf{p}}_k^{\ell-1} & \text{if } \ell \ge J
\end{cases}
\end{equation}
Here, $\mathbf{p}_k^{\text{in}, \ell}$ is the input prompt for layer $\ell$, $\tilde{\mathbf{p}}_k^{\ell-1}$ is the context-aware output from the previous layer, and $\mathbf{p}_k$ is the initial raw token. This controlled mixture, governed by $\beta$, allows contextual adaptation while preventing semantic drift. At inference, only the final [EOT] feature is used.
\begin{figure}[t!] 
    \centering 
    \includegraphics[width=0.5\textwidth]{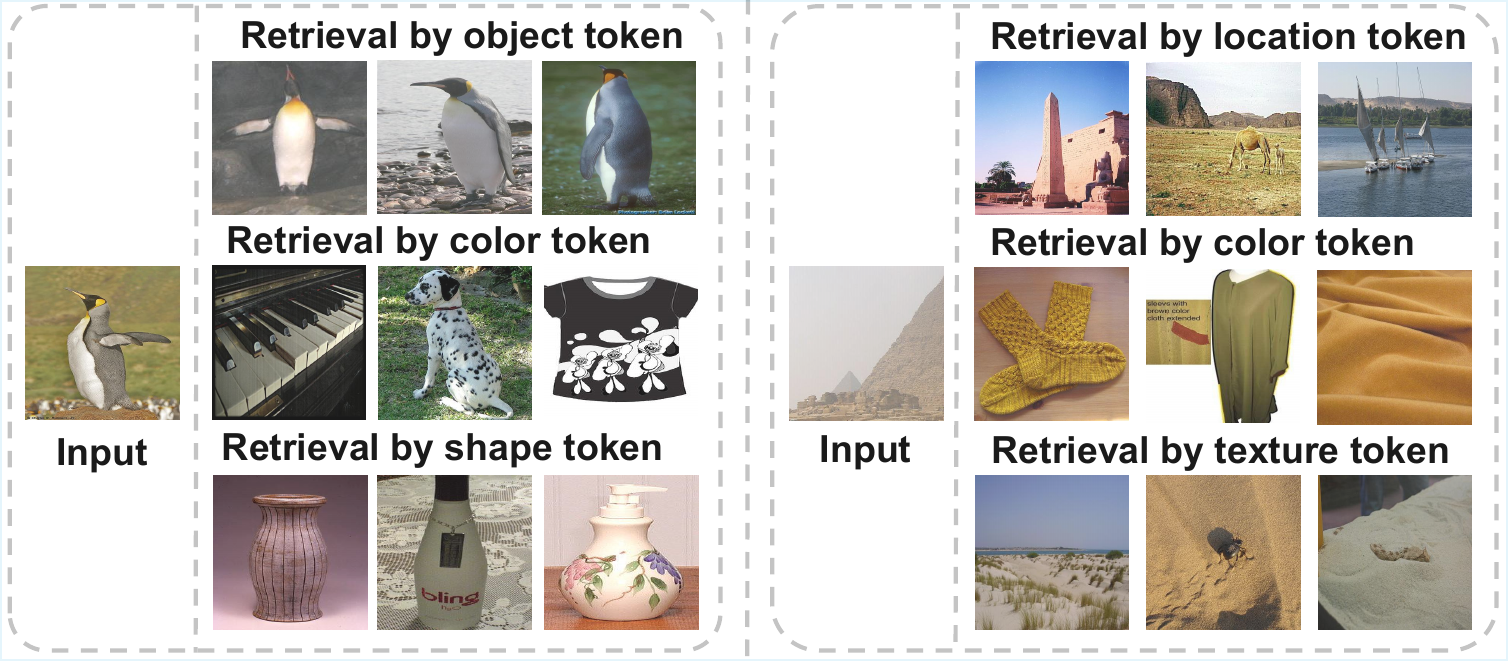} 
    \caption{
        Results of attribute-conditioned image retrieval on ImageNet. For each query image, we retrieval using features from different learned attribute tokens.
    }
    \label{fig:retrieval_analysis} 
\end{figure}

\subsubsection{Role-Based Attention Mask}
\label{sec:role_based_mask} 

Preserving generalization requires precise control over the information flow between the original and newly introduced attribute tokens. Building on our data-driven identification of head roles, we introduce a \textbf{role-based attention mask} mechanism. This mechanism, applied within the deeper layers of the ViT ($l \ge J$), assigns a custom attention mask $\mathbf{M}_{(l,h)} \in \{-\infty, 0\}^{N_{seq} \times N_{seq}}$ to each head $(l,h)$ based on its specific function. The mask is applied additively to the raw attention scores:
\begin{equation}
\mathbf{S}_{(l,h)} = \frac{\mathbf{Q}_{(l,h)}\mathbf{K}_{(l,h)}^T}{\sqrt{D_h}} + \mathbf{M}_{(l,h)}
\label{eq:masked_scores}
\end{equation}
An entry $\mathbf{M}_{(l,h)}[i, j] = -\infty$ effectively severs the connection from query token $i$ to key token $j$ by setting its corresponding attention weight to zero post-softmax.

To formally define the masks, we partition the token indices into disjoint sets: $\mathcal{I}_{\text{cls}}$ for the \texttt{[CLS]} token, $\mathcal{I}_{\text{patch}}$ for patch tokens, and $\mathcal{I}_{\text{attr}}$ for our learnable attribute tokens. The set of original token indices is $\mathcal{I}_{\text{orig}} = \mathcal{I}_{\text{cls}} \cup \mathcal{I}_{\text{patch}}$. The mask construction depends on the head's role:

\paragraph{Generalization and Other Specialized Heads.}
This category includes both \textbf{Generalization Heads}, and other \textbf{Specialized Heads} not selected as core attributes (e.g., for 'numbers' or 'text'). We hypothesize that both types of heads encode the model's intrinsic, pre-trained capabilities that are crucial for zero-shot generalization. To protect these foundational abilities from being perturbed by task-specific information, we implement a strict mask. This mask prohibits any interaction between the original tokens and our five learnable attribute tokens:
\begin{equation}
\mathbf{M}_{(l,h)}[i, j] = 
\begin{cases} 
-\infty & \text{if } (i \in \mathcal{I}_{\text{orig}} \text{ and } j \in \mathcal{I}_{\text{attr}}) \\
& \text{or } (i \in \mathcal{I}_{\text{attr}} \text{ and } j \in \mathcal{I}_{\text{orig}}) \\
0 & \text{otherwise}
\end{cases}
\end{equation}
This policy ensures that both generalist heads and other specialist heads remain isolated, preserving their original functions.

\paragraph{Core Attribute Heads.}
These are the heads identified as specializing in one of our five core visual attributes $a \in A$ (\textit{e.g., 'color'}). The mask for such a head is designed to channel the corresponding learnable attribute token exclusively to its expert head, enabling focused learning. Let $\mathcal{I}_{\text{attr}(a)} \subset \mathcal{I}_{\text{attr}}$ be the index set for the token(s) corresponding to attribute $a$. The mask is defined as:
\begin{equation}
\mathbf{M}_{(l,h)}[i, j] = 
\begin{cases} 
-\infty & \text{if } j \in \mathcal{I}_{\text{attr}} \text{ and } j \notin \mathcal{I}_{\text{attr}(a)} \\
0 & \text{otherwise}
\end{cases}
\end{equation}
This rule allows any query token to attend to all original tokens and its relevant attribute token, while blocking attention to irrelevant ones. This promotes clean, disentangled attribute learning.

\paragraph{Mixed Heads.}
Heads with intermediate Concept Entropy are classified as \textbf{Mixed Heads}. We consider their function to be multifaceted and flexible. Therefore, we permit unrestricted attention by applying an all-zeros mask where $\mathbf{M}_{(l,h)}[i, j] = 0, \forall i, j$, allowing them to integrate information freely from all available tokens.

\subsubsection{Final Feature Generation.}
After passing through all $L$ layers, the final hidden states of the \texttt{[CLS]} and attribute tokens are projected to generate features. The state of the original class token, $\mathbf{z}_{\text{cls}}^L$, is projected by the pre-trained and \textbf{frozen} vision projection head, $\mathbf{P}_{\text{vision}}(\cdot)$, to yield the class feature $\mathbf{f}_{\text{cls}}$:
\begin{equation}
    \label{eq:fc_generation}
    \mathbf{f}_{\text{cls}} = \mathbf{P}_{\text{vision}}(\text{LN}(\mathbf{z}_{\text{cls}}^L))
\end{equation}
Similarly, the final state of each attribute token, $\mathbf{r}_{\text{attr}}^L$, is processed by a shared, \textbf{trainable} projection head, $\mathbf{P}_{\text{attr}}(\cdot)$, to yield a set of specialized attribute features $\{\mathbf{f}_{\text{attr}}\}$:
\begin{equation}
    \label{eq:fr_generation}
    \mathbf{f}_{\text{attr}} = \mathbf{P}_{\text{attr}}(\text{LN}(\mathbf{r}_{\text{attr}}^L))
\end{equation}
This process yields a primary feature $\mathbf{f}_cls$ retaining generalization, and specialized features $\mathbf{f}_\text{attr}$.


To visually validate our attribute tokens capture semantic concepts, we conduct an attribute-conditioned image retrieval, shown in Figure~\ref{fig:retrieval_analysis}.

\subsubsection{Training and Inference}
\label{sec:training_inference}

\paragraph{Training.}
DeAR's training process balances classification accuracy on the target task with preserving the model's pre-trained generalization. The total loss function is a weighted combination of a primary classification loss and two regularization terms:
\begin{equation}
    \mathcal{L}_{\text{total}} = \mathcal{L}_{\text{CE}} + \lambda_{\text{reg}} \mathcal{L}_{\text{reg}} + \lambda_{\text{fusion}} \mathcal{L}_{\text{fusion}}
\label{eq:total_loss}
\end{equation}
where $\lambda_{\text{reg}}$ and $\lambda_{\text{fusion}}$ are hyperparameters that balance the contribution of each component.

The primary loss, $\mathcal{L}_{\text{CE}}(\hat{\mathbf{y}}, \mathbf{y})$, is the standard cross-entropy between the predicted logits $\hat{\mathbf{y}}$ (defined in our inference strategy \citep{mao2023crossentropylossfunctionstheoretical}, Eq.~\ref{eq:fusion_explicit}) and the ground-truth labels $\mathbf{y}$.

To preserve the model's foundational knowledge, we introduce a self-regularization loss, $\mathcal{L}_{\text{reg}}$. This term encourages the final features from both the vision and text encoders to remain close to their counterparts from the original, frozen CLIP model. Specifically, we regularize the final class feature $\mathbf{f}_{\text{cls}}$ and the text feature $\mathbf{f}_t$:
\begin{align}
    \mathcal{L}_{\text{reg}} &= \mathcal{L}_{\text{reg}}^V + \mathcal{L}_{\text{reg}}^T \\
    \quad \mathcal{L}_{\text{reg}}^V &= 1 - \text{cos}(\mathbf{f}_{\text{cls}}, \mathbf{f}_{\text{cls}}^{\text{orig}}) \\
    \quad \mathcal{L}_{\text{reg}}^T &= 1 - \text{cos}(\mathbf{f}_t, \mathbf{f}_{t}^{\text{orig}})
\end{align}
Here, $\mathbf{f}_{\text{cls}}^{\text{orig}}$ and $\mathbf{f}_{t}^{\text{orig}}$ are the features extracted from the unmodified CLIP model, and $\text{cos}(\cdot, \cdot)$ denotes the cosine similarity.

To prevent over-reliance on the newly learned attribute features, we introduce a fusion-weight regularization, $\mathcal{L}_{\text{fusion}}$. This loss encourages the inference mechanism to assign a higher weight to the more general class feature:
\begin{equation}
    \mathcal{L}_{\text{fusion}} = -\log(\alpha_{\text{cls}})
\end{equation}
where $\alpha_{\text{cls}}$ is the softmax-normalized weight corresponding to the class feature.

\paragraph{Task-Adaptive Fusion for Inference.}
For inference, we propose a \textbf{Task-Adaptive Fusion} strategy that combines evidence from the protected class and specialized attribute features. We learn a small set of scalar weights, $\{w_k\}_{k \in \{\text{cls}\} \cup A}$, on the base classes, which are then fixed for novel classes.

For a given image and $m$ class text embeddings $\{\mathbf{t}_j\}_{j=1}^m$, we first compute similarity vectors for the class feature $\mathbf{f}_{\text{cls}}$ and each attribute feature $\mathbf{f}_{a}$:
\begin{align}
    \mathbf{s}_{\text{cls}} &= \tau \cdot [\text{sim}(\mathbf{f}_{\text{cls}}, \mathbf{t}_1), \dots, \text{sim}(\mathbf{f}_{\text{cls}}, \mathbf{t}_m)] \\
    \mathbf{s}_{a} &= \tau \cdot [\text{sim}(\mathbf{f}_{a}, \mathbf{t}_1), \dots, \text{sim}(\mathbf{f}_{a}, \mathbf{t}_m)]
\end{align}
where $\tau$ is CLIP's frozen temperature parameter. The final logits $\hat{\mathbf{y}}$ are a weighted sum of these vectors. The fusion weights $\alpha_k$ are derived from the learned scalars $\{w_k\}$ via softmax, where $\alpha_k = e^{w_k} / (e^{w_{\text{cls}}} + \sum_{i \in A} e^{w_i})$. The final logits are then computed as:
\begin{equation}
    \hat{\mathbf{y}} = \alpha_{\text{cls}} \cdot \mathbf{s}_{\text{cls}} + \sum_{a \in A} \alpha_a \cdot \mathbf{s}_a.
    \label{eq:fusion_explicit}
\end{equation}
This strategy assumes that a model's reliance on specific attributes is a task-level prior (e.g., color is important for bird classification). By learning and transferring these priors, we achieve a more principled and effective fusion.


\section{Experiments}
\begin{table*}[ht]
\centering
\caption{Comparison of DeAR with previous methods on base-to-novel generalization across 11 datasets. Bold values indicate the best results.}
\resizebox{\textwidth}{!}{
\begin{tabular}{l|ccc|ccc|ccc|ccc|ccc|ccc}
\toprule
\multirow{2}{*}{Method} 
    & \multicolumn{3}{c|}{Average} 
    & \multicolumn{3}{c|}{ImageNet} 
    & \multicolumn{3}{c|}{Caltech101} 
    & \multicolumn{3}{c|}{OxfordPets}
    & \multicolumn{3}{c|}{StanfordCars} 
    & \multicolumn{3}{c}{Flowers102} \\
 & Base & Novel & HM & Base & Novel & HM & Base & Novel & HM & Base & Novel & HM & Base & Novel & HM & Base & Novel & HM \\
\midrule
CLIP \scriptsize{(ICML2021)}          & 69.34 & 74.22 & 71.70 & 72.43 & 68.14 & 70.22 & 96.84 & 94.00 & 95.40 & 91.17 & 97.96 & 94.12 & 63.37 & 74.89 & 68.65 & 72.08 & 77.80 & 74.83 \\
CoOp \scriptsize{(ICCV2021)}          & 82.69 & 63.22 & 71.66 & 76.47 & 67.88 & 71.92 & 98.01 & 89.81 & 93.73 & 93.67 & 95.29 & 94.47 & 78.12 & 60.40 & 68.13 & 97.60 & 59.67 & 74.06 \\
ATPrompt$_{CoOp}$ \scriptsize{(ICCV2025)} & 82.68 & 68.04 & 74.65 & 76.27 & 70.60 & 73.33 & 97.95 & 93.63 & 95.74 & 94.77 & 96.59 & 95.67 & 77.43 & 66.55 & 71.58 & 97.44 & 67.52 & 79.77 \\
ProGrad \scriptsize{(CVPR2023)}         & 82.48 & 70.75 & 76.16 & 77.02 & 66.66 & 71.46 & 98.02 & 93.89 & 95.91 & 95.07 & 97.63 & 96.33 & 77.68 & 68.63 & 72.88 & 95.54 & 71.87 & 82.03 \\
MaPLe \scriptsize{(CVPR2023)}         & 82.28 & 75.14 & 78.55 & 76.60 & 70.54 & 73.47 & 97.74 & 94.36 & 96.02 & 95.43 & 97.76 & 96.58 & 72.94 & 74.00 & 73.47 & 95.92 & 72.46 & 82.56 \\
PromptSRC \scriptsize{(ICCV2023)}     & 84.26 & 76.10 & 79.97 & 77.60 & 70.73 & 74.01 & 98.10 & 94.03 & 96.02 & 95.33 & 97.30 & 96.30 & 78.27 & 74.97 & 76.58 & 98.07 & 76.50 & 85.95 \\
ProVP \scriptsize{(CVPR2024)}        & 85.20 & 73.22 & 78.76 & 75.82 & 69.21 & 72.36 & 98.92 & 94.21 & 96.51 & 95.87 & 97.65 & 96.75 & 80.43 & 67.96 & 73.67 & 98.42 & 72.06 & 83.20 \\
TCP \scriptsize{(CVPR2024)}           & 84.13 & 75.36 & 79.51 & 77.27 & 69.87 & 73.38 & 98.23 & 94.67 & 96.42 & 94.67 & 97.20 & 95.92 & 80.80 & 74.13 & 77.32 & 97.73 & 75.57 & 85.23 \\
PromptKD \scriptsize{(CVPR2024)}      & 84.11 & 78.28 & 81.09 & 77.63 & 70.96 & 74.15 & 98.31 & \textbf{96.29} & \textbf{97.29} & 93.42 & 97.44 & 95.39 & 80.48 & \textbf{81.78} & \textbf{81.12} & 98.69 & \textbf{81.91} & \textbf{89.52} \\
MMRL \scriptsize{(CVPR2025)}          & 85.68 & 77.16 & 81.20 & 77.90 & 71.30 & 74.45 & 98.97 & 94.50 & 96.68 & 95.90 & 97.60 & 96.74 & 81.30 & 75.07 & 78.06 & 98.97 & 77.27 & 86.78 \\
TAC \scriptsize{(CVPR2025)}           & 85.24 & 77.60 & 81.24 & \textbf{78.57} & 71.03 & 74.61 & 98.57 & 95.27 & 96.89 & 95.93 & \textbf{98.17} & \textbf{97.04} & 81.63 & 74.17 & 77.72 & 97.97 & 76.87 & 86.15 \\
SkipT \scriptsize{(CVPR2025)} & 85.04 & 77.53 & 81.11 & 77.73 & 70.40 & 73.89 & 98.50 & 95.33 & 96.89 & 95.70 & 97.87 & 96.77 & \textbf{82.93} & 72.50 & 77.37 & 98.57 & 75.80 & 85.70 \\
\midrule
\rowcolor{lightgray} 
\textbf{DeAR (Ours)}                   & \textbf{85.94} & \textbf{79.73} & \textbf{82.72} & 78.12 & \textbf{71.80} & \textbf{74.83} & \textbf{99.00} & 95.80 & 96.95 & \textbf{97.34} & 97.50 & 96.81 & 82.01 & 75.20 & 78.47 & \textbf{99.00} & 78.50 & 87.57 \\
\bottomrule
\end{tabular}
}
\vspace{0.1cm}

\resizebox{\textwidth}{!}{
\begin{tabular}{l|ccc|ccc|ccc|ccc|ccc|ccc}
\toprule
\multirow{2}{*}{Method} 
    & \multicolumn{3}{c|}{Food101} 
    & \multicolumn{3}{c|}{FGVC Aircraft} 
    & \multicolumn{3}{c|}{SUN397} 
    & \multicolumn{3}{c|}{DTD}
    & \multicolumn{3}{c|}{EuroSAT}
    & \multicolumn{3}{c}{UCF101} \\
 & Base & Novel & HM & Base & Novel & HM & Base & Novel & HM & Base & Novel & HM & Base & Novel & HM & Base & Novel & HM \\
\midrule
CLIP \scriptsize{(ICML2021)}          & 90.10 & 91.22 & 90.66 & 27.19 & 36.29 & 31.09 & 69.36 & 75.35 & 72.23 & 53.24 & 59.90 & 56.37 & 56.48 & 64.05 & 60.03 & 70.53 & 77.50 & 73.85 \\
CoOp \scriptsize{(ICCV2021)}          & 88.33 & 82.26 & 85.19 & 40.44 & 22.30 & 28.75 & 80.60 & 65.89 & 72.51 & 79.44 & 41.18 & 54.24 & 92.19 & 54.74 & 68.69 & 84.69 & 56.05 & 67.46 \\
ATPrompt$_{CoOp}$ \scriptsize{(ICCV2025)} & 88.74 & 87.44 & 88.09 & 40.38 & 27.22 & 32.52 & 80.84 & 68.64 & 74.24 & 80.83 & 45.49 & 58.22 & 90.34 & 59.79 & 71.96 & 84.49 & 64.96 & 73.45 \\
ProGrad \scriptsize{(CVPR2023)}         & 90.37 & 89.59 & 89.98 & 40.54 & 27.57 & 32.82 & 81.26 & 74.17 & 77.55 & 77.35 & 52.35 & 62.45 & 90.11 & 60.89 & 72.67 & 84.33 & 74.94 & 79.35 \\
MaPLe \scriptsize{(CVPR2023)}         & 90.71 & 92.05 & 91.38 & 37.44 & 35.61 & 36.50 & 80.82 & 78.70 & 79.75 & 80.36 & 59.18 & 68.16 & 94.07 & 73.23 & 82.35 & 83.00 & 78.66 & 80.77 \\
PromptSRC \scriptsize{(ICCV2023)}     & 90.67 & 91.53 & 91.10 & 42.73 & 37.87 & 40.15 & 82.67 & 78.47 & 80.52 & 83.37 & 62.97 & 71.75 & 92.90 & 73.90 & 82.32 & 87.10 & 78.80 & 82.74 \\
ProVP \scriptsize{(CVPR2024)}        & 90.32 & 90.91 & 90.61 & 47.08 & 29.87 & 36.55 & 83.14 & 74.71 & 78.73 & 62.11 & 65.53 & 63.77 & \textbf{97.12} & 72.91 & 83.29 & 87.30 & 79.00 & 83.00 \\
TCP \scriptsize{(CVPR2024)}           & 90.57 & 91.37 & 90.97 & 41.97 & 34.43 & 37.83 & 82.63 & 78.20 & 80.35 & 82.77 & 58.07 & 68.25 & 91.63 & 74.73 & 82.32 & 87.13 & 80.77 & 83.83 \\
PromptKD \scriptsize{(CVPR2024)}      & 89.43 & 91.27 & 90.34 & 43.61 & \textbf{39.68} & 41.55 & 82.53 & 80.88 & 81.70 & 82.86 & 69.15 & 75.39 & 92.04 & 71.59 & 80.54 & 86.23 & 80.11 & 83.06 \\
MMRL \scriptsize{(CVPR2025)}          & 90.57 & 91.50 & 91.03 & 46.30 & 37.03 & 41.15 & 83.20 & 79.30 & 81.20 & \textbf{85.67} & 65.00 & 73.82 & 95.60 & 80.17 & 87.21 & \textbf{88.10} & 80.07 & 83.89 \\
TAC \scriptsize{(CVPR2025)}           & \textbf{90.87} & 91.87 & \textbf{91.37} & 44.60 & 37.70 & 40.86 & 83.70 & 80.03 & 81.82 & 83.37 & 64.27 & 72.58 & 94.37 & 82.60 & 88.10 & 88.07 & 81.67 & 84.75 \\
SkipT \scriptsize{(CVPR2025)} & 90.67 & \textbf{92.03} & 91.34 & 45.37 & 37.13 & 40.84 & 82.40 & 79.03 & 80.68 & 83.77 & 67.23 & 74.59 & 92.47 & 83.00 & 87.48 & 87.30 & \textbf{82.47} & 84.81 \\
\midrule
\rowcolor{lightgray} 
\textbf{DeAR (Ours)}                       & 90.10 & 91.90 & 90.93 & \textbf{47.10} & 38.90 & \textbf{42.61} & \textbf{83.82} & \textbf{81.80} & \textbf{82.80} & 83.90 & \textbf{75.60} & \textbf{79.62} & 97.00 & \textbf{87.90} & \textbf{92.22} & 87.90 & 82.10 & \textbf{84.90} \\
\bottomrule
\end{tabular}
}
\label{b2n}
\end{table*}

\subsection{Experiment settings}
\paragraph{Implementation Details.}
Our implementation is based on the OpenCLIP with a ViT-B-16 backbone. We train the model using the AdamW optimizer with an initial learning rate of $1 \times 10^{-3}$. All experiments are conducted on three NVIDIA RTX 4090 GPUs. Further details are provided in the Appendix.
\paragraph{Base-to-Novel Generalization.}
This is the primary evaluation to assess DeAR's ability to adapt to downstream tasks while preserving generalization. We conduct this evaluation across eleven image classification datasets: ImageNet \citep{5206848}, Caltech101 \citep{1384978}, OxfordPets \citep{6248092}, StanfordCars \citep{6755945}, Flowers102 \citep{4756141}, Food101 \citep{10.1007/978-3-319-10599-4_29}, FGVCAircraft \citep{maji2013finegrainedvisualclassificationaircraft}, SUN397 \citep{5539970}, UCF101 \citep{Soomro2012UCF101AD}, DTD \citep{6909856}, and EuroSAT \citep{8736785}. We follow the standard setup: the model trains on 16-shot base classes and is evaluated on the test sets of both base and novel classes. As shown in Table~\ref{b2n}, DeAR demonstrates clear superiority, achieving a new state-of-the-art harmonic mean of 82.72\%. This result is driven by a significant 1.83\% improvement on novel classes over the previous best, MMRL. The substantial gain on unseen classes directly validates our core hypothesis: protecting generalization heads is crucial for preserving the model's foundational knowledge.

\paragraph{Domain Generalization.}
This evaluation measures the model's robustness to out-of-distribution data and domain shifts. Following standard practice, we train the model on the 16-shot ImageNet training set (all classes) and directly evaluate its performance, without further fine-tuning, on four challenging ImageNet variants: ImageNetV2 \citep{Recht2019DoIC}, ImageNet-Sketch \citep{NEURIPS2019_3eefceb8}, ImageNet-A \citep{9578772}, and ImageNet-R \citep{9710159}. Table~\ref{tab:domain_gen_results} presents the domain generalization results. Our method demonstrates consistently strong robustness across all four out-of-distribution datasets. DeAR achieves the best performance on ImageNet-A (51.80\%) and ImageNet-R (78.83\%), also is highly competitive in the other variants.

\begin{table*}[ht]
\centering
\caption{Comparison of DeAR with previous methods on cross-dataset evaluation. Bold values indicate the best results.}
\resizebox{\textwidth}{!}{
\begin{tabular}{llc|ccccccccccc}
\toprule
& & \multicolumn{1}{c|}{Source} & \multicolumn{11}{c}{Target} \\
\cmidrule(lr){3-3} \cmidrule(lr){4-14}
 &  & {ImageNet} & {Average} & {Caltech101} & {OxfordPets} & {StanfordCars} & {Flowers102} & {Food101} & {FGVC-\newline Aircraft} & {SUN397} & {DTD} & {EuroSAT} & {UCF101} \\
\midrule
CoOp       & & 71.51 & 63.88 & 93.70 & 89.14 & 64.51 & 68.71 & 85.30 & 18.47 & 64.15 & 41.92 & 46.39 & 66.55 \\
MaPLe      &  & 70.72 & 66.30 & 93.53 & 90.49 & 65.57 & 72.23 & 86.20 & 24.74 & 67.01 & 46.49 & 48.06 & 68.69 \\
PromptSRC  &  & 71.27 & 65.81 & 93.60 & 90.25 & 65.70 & 70.25 & 86.15 & 23.90 & 67.10 & 46.87 & 45.50 & 68.75 \\
MMRL   & & 72.03 & 67.25 & 94.67 & \textbf{91.43} & 66.10 & 72.77 & 86.40 & 26.30 & 67.57 & 45.90 & 53.10 & 68.27 \\
TAC & & \textbf{72.77} & 66.53 & 94.53 & 90.67 & 65.30 & 72.20 & 85.83 & 23.53 & 67.63 & \textbf{47.57} & 48.07 & \textbf{70.00} \\
\midrule
\rowcolor{lightgray} 
\textbf{DeAR (Ours)}   & & 71.50 & \textbf{67.60} & \textbf{95.00} & 91.23 & \textbf{66.70} & \textbf{73.80} & \textbf{86.80} & \textbf{26.50} & \textbf{68.30} & 45.40 & \textbf{53.50} & 68.80 \\
\bottomrule
\end{tabular}
}
\label{tab:cross_dataset_results}
\end{table*}

\begin{figure}[htbp]
     \centering
     \includegraphics[width=0.4\textwidth]{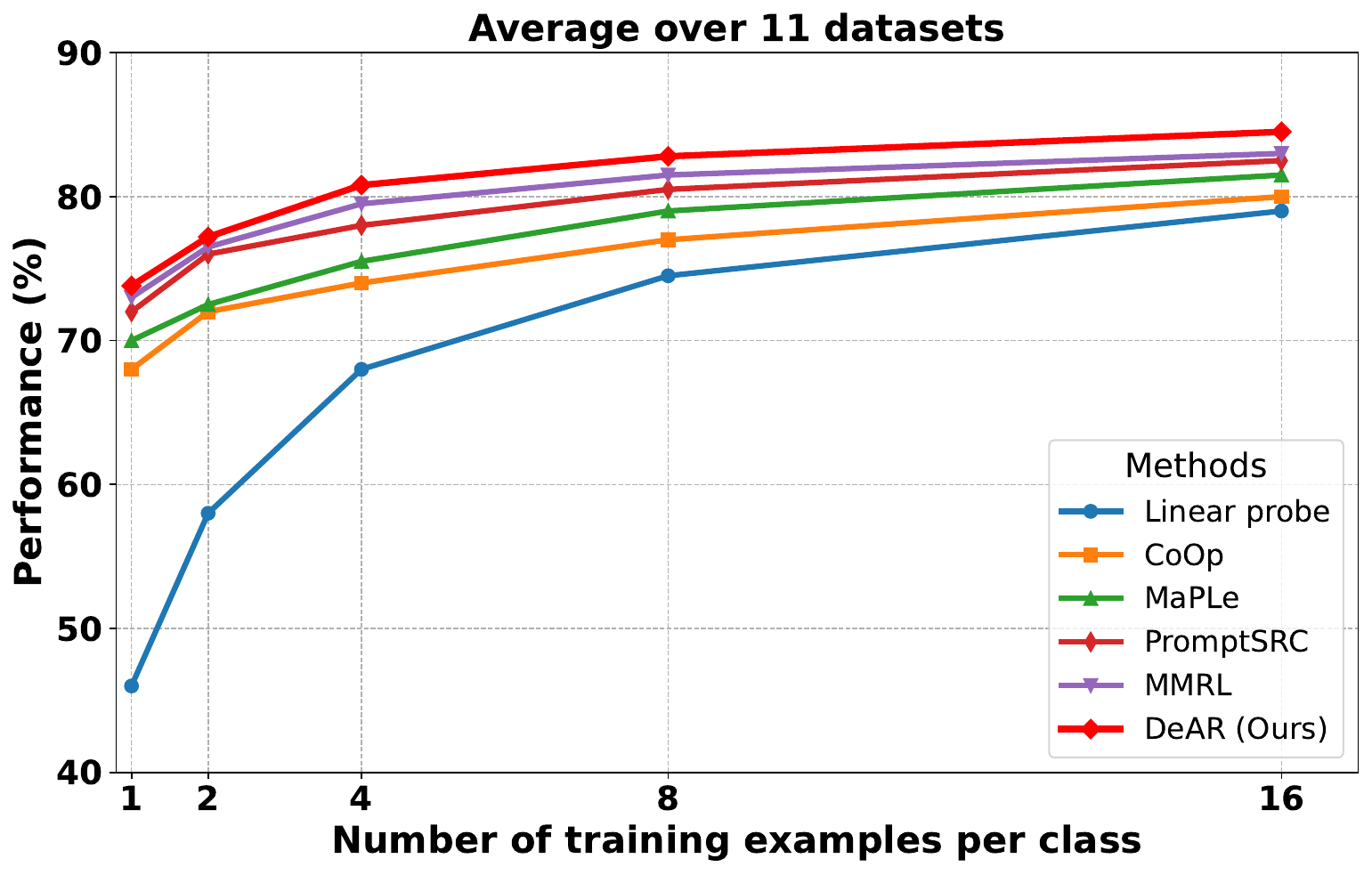}%
    \caption{Comparison of DeAR with previous methods on few-shot learning. Detailed results are provided in the Appendix}
    \label{fig:few_shot}
\end{figure}

\begin{table}[ht]
\centering
\caption{Comparison of DeAR with previous methods on domain generalization. Bold values indicate the best results.}
\resizebox{0.5\textwidth}{!}{
\begin{tabular}{llc|ccccc}
\toprule
& & \multicolumn{1}{c|}{Source} & \multicolumn{5}{c}{Target} \\
\cmidrule(lr){3-3} \cmidrule(lr){4-8}
 &  & ImageNet & Average & -V2 & -S & -A & -R \\
\midrule
CoOp        & & 71.51 & 59.28 & 64.20 & 47.99 & 49.71 & 75.21 \\
MaPLe       & & 70.72 & 60.28 & 64.07 & 49.15 & 50.90 & 76.98 \\
PromptSRC   & & 71.27 & 60.65 & 64.35 & 49.55 & 50.90 & 77.80 \\
MMRL    & & 72.03 & 60.59 & 64.47 & 49.17 & 51.20 & 77.53 \\
TAC & & \textbf{72.77} & \textbf{61.63} & \textbf{65.97} & \textbf{50.30} & 51.73 & 78.50 \\
\midrule
\rowcolor{lightgray} 
\textbf{DeAR (Ours)}  & & 71.50 & 61.25 & 64.87 & 49.50 & \textbf{51.80} & \textbf{78.83} \\
\bottomrule
\end{tabular}
}
\label{tab:domain_gen_results}
\end{table}
\paragraph{Few-Shot Learning.}
Few-shot learning evaluates the capacity of the model for knowledge transfer and generalization when labeled data are limited. To quantify this capability, we fine-tune DeAR on $k$ shot subsets of the original training data, where $k\in\{1,2,4,8,16\}$. As shown in Figure~\ref{fig:few_shot}, across all settings from 1 to 16 shots, DeAR consistently outperforms the baseline methods, demonstrating a slight but consistent performance advantage. This suggests that our DeAR framework provides robust performance even in limited labeled data. Detailed results on all 11 datasets are provided in the Appendix. 

\paragraph{Cross-Dataset Generalization.}
To evaluate transfer ability, we train DeAR on ImageNet (16 shots per class) and test it zero-shot on ten unseen datasets. Since the Task-Adaptive Fusion weights are optimized for ImageNet, they may not generalize. Therefore, for a fair evaluation on the target datasets, we adopt a decoupled inference strategy similar to MMRL~\citep{guo2025mmrl}, using only the protected class feature $\mathbf{f}_{\text{cls}}$ for prediction. As shown in Table~\ref{tab:cross_dataset_results}, DeAR sets a new state-of-the-art average accuracy of 67.60\%, surpassing the previous best method, MMRL, while maintaining comparable performance on the source task (ImageNet).

\subsection{Ablation Studies}
To validate the effectiveness of the key components in our DeAR framework, we conduct a series of ablation studies on the average performance across all 11 datasets in the base-to-novel generalization setting.

\paragraph{Role-Based Attention Mask.}
To validate our core hypothesis, that role-based information control is superior to uniform strategies, we compare DeAR against two variants: All-Generalization, where all deep heads are shielded from attribute tokens, and All-Mixed, where all interactions are permitted.


The results shown in Table~\ref{tab:ablation_Mask} validate our role-based design. The All-Generalization strategy maximizes novel-class performance at the cost of base-class learning, while the All-Mixed strategy does the opposite. Our proposed DeAR strikes the optimal balance between adaptation and generalization, achieving the best overall Harmonic Mean.

\begin{table}[h!]
\centering
\caption{Ablation on Role-Based Attention Mask.}
\small 
\begin{tabular}{lccc}
    \toprule
    Mask Strategy & Base & Novel & HM \\
    \midrule
    All-Generalization & 81.25 & \textbf{80.12} & 80.68 \\
    All-Mixed & 85.31 & 76.98 & 80.93 \\
    \rowcolor{gray!20} 
    \textbf{DeAR (Ours)} & \textbf{85.94} & 79.73 & \textbf{82.72} \\
    \bottomrule
\end{tabular}
\label{tab:ablation_Mask}
\end{table}

\paragraph{Impact of Loss Components.}
We also analyze the contribution of each component in our proposed loss function. We start with a baseline using only the cross-entropy classification loss ($\mathcal{L}_{\text{CE}}$) and incrementally add our two regularization terms: the multimodal self-regularization ($\mathcal{L}_{\text{reg}}$) and the fusion-weight regularization ($\mathcal{L}_{\text{fusion}}$).


As shown in Table~\ref{tab:ablation_loss}, each loss component is vital. Adding $\mathcal{L}_{\text{reg}}$ significantly boosts Novel class performance (+4.24\%). The subsequent addition of $\mathcal{L}_{\text{fusion}}$ further improves the overall Harmonic Mean, demonstrating the benefit of prioritizing the class feature.

\begin{table}[h!]
\centering
\caption{Ablation on Loss function components.}
\small 
\begin{tabular}{lccc}
    \toprule
    Loss Function & Base & Novel & HM \\
    \midrule
    $\mathcal{L}_{\text{CE}}$ & \textbf{86.75} & 74.21 & 79.99 \\
    $\mathcal{L}_{\text{CE}}$ + $\mathcal{L}_{\text{reg}}$ & 85.03 & 78.45 & 81.61 \\
    \rowcolor{gray!20} 
    $\mathcal{L}_{\text{CE}}$ + $\mathcal{L}_{\text{reg}}$ + $\mathcal{L}_{\text{fusion}}$ & 85.94 & \textbf{79.73} & \textbf{82.72} \\
    \bottomrule
\end{tabular}
\label{tab:ablation_loss}
\end{table}

\section{Conclusion}

In this work, we propose the DeAR framework to enhance the adaptation of CLIP by balancing task-specific knowledge with generalization preservation. DeAR offers a fine-grained controllable fine-tuning process. Using our Concept Entropy, we decompose each attention head roles, allowing DeAR to direct new knowledge to attribute heads while protecting generalization heads. Experiments show that DeAR achieves state-of-the-art performance on base-to-novel generalization benchmarks and performs well on various tasks. More importantly, the learned attribute-aware representations offer significant potential for future applications requiring explicit semantic control, such as fine-grained retrieval. We demonstrate that fine-grained internal control is the path to better VLMs adaptation.
{
    \small
    \bibliographystyle{ieeenat_fullname}
    \bibliography{main}
}


\end{document}


\maketitle

\appendix
\section{Implementation Details}
\label{sec:appendix_implementation}

This section provides a comprehensive overview of the implementation details used for all experiments to ensure full reproducibility. Our code is built upon PyTorch and will be made publicly available.

\subsection{Dataset and Dataloader Configuration}
For all datasets, we follow a consistent data loading and augmentation pipeline:
\begin{itemize}[leftmargin=*]
    \item \textbf{Image Pre-processing:} Input images are resized to $224 \times 224$ pixels. For training, we apply standard augmentations including \texttt{random\_resized\_crop} with a scale range of $(0.5, 1.0)$ and \texttt{random\_flip}. The interpolation method used is bicubic. Finally, all images are normalized using the standard CLIP pixel mean $[0.4814, 0.4578, 0.4082]$ and standard deviation $[0.2686, 0.2613, 0.2757]$.
    \item \textbf{Dataloader:} For training, we use a batch size of 16. For evaluation, the batch size is set to 250. We use 8 worker threads for data loading to maximize efficiency.
\end{itemize}

\subsection{Training and Optimization}
Our training setup is designed for stable and efficient convergence:
\begin{itemize}[leftmargin=*]
    \item \textbf{Optimizer:} We use the AdamW optimizer for all learnable parameters.
    \item \textbf{Learning Rate Schedule:} The initial learning rate is set to $1 \times 10^{-3}$. We employ a cosine annealing learning rate scheduler over a total of 10 epochs.
    \item \textbf{Warmup:} A 1-epoch constant warmup period is used at the beginning of training, with a warmup learning rate of $1 \times 10^{-5}$. This helps stabilize the model in the early stages of training.
    \item \textbf{Training Precision:} We use Automatic Mixed Precision (AMP) to accelerate training and reduce memory consumption.
\end{itemize}

\subsection{Model and DeAR Configuration}
\paragraph{Backbone and Layer Selection.} All experiments are conducted using the official pre-trained CLIP model with a ViT-B/16 backbone. We strategically inject our learnable tokens into the deep layers (Layers 9 to 12) of the encoders, corresponding to $J=9$ in the main text. This design choice is guided by prior findings demonstrating that functional specialization in Vision Transformers predominantly emerges in the final few layers. We further validated this through mean-ablation on our specific ViT-B/16 backbone.

\paragraph{Vision-Text Asymmetry.} While the update mechanism is symmetric across modalities, the token definitions differ by design. Both the vision and text encoders insert learnable tokens at layers $L_{9-12}$ with the same $\beta$-controlled flow (Eq. 5 \& 6 in the main text) to prevent semantic drift. However, we intentionally exclude explicit Attribute Tokens from the text side. In CLIP-based classification, text features act as fixed anchors (via templates like ``a photo of a \{class\}'') for semantic alignment. Keeping the text side simplified prevents over-complication while maintaining the integrity of these semantic anchors.

\paragraph{DeAR Hyperparameters.}
\begin{itemize}[leftmargin=*]
    \item \textbf{Token Injection:} We use a learnable prompt/token length of $K=5$, selected via Density-Based Clustering Validation (DBCV).
    \item \textbf{Context Information Control:} The hyperparameter $\beta$ which controls the context information flow is set to 0.9.
    \item \textbf{Loss Weights:} The balancing hyperparameters for our composite loss function are set to $\lambda_{\text{reg}}=1.0$ and $\lambda_{\text{fusion}}=0.7$. All hyperparameters were fixed across all 11 datasets after initial tuning on a validation set.
\end{itemize}

\paragraph{Computational Cost.} 
During training, the regularization term $\mathcal{L}_{\text{reg}}$ requires one additional forward pass through the frozen backbone, which increases the overall training cost by approximately $35\%$. We find this trade-off highly acceptable given the significant zero-shot generalization gains achieved. Crucially, during inference, the computational overhead is exactly zero, as the final adapted model retains the standard ViT architecture without requiring auxiliary modules. All experiments were conducted on a server equipped with three NVIDIA RTX 4090 GPUs.

\begin{table}[h!]
\centering
\caption{Summary of key hyperparameters used in our experiments.}
\label{tab:hyperparams}
\begin{tabular}{lc}
\toprule
\textbf{Hyperparameter} & \textbf{Value} \\
\midrule
\multicolumn{2}{l}{\textit{Dataloader \& Input}} \\
Batch Size (Train / Test) & 16 / 250 \\
Input Resolution & $224 \times 224$ \\
Augmentations & \texttt{RRCrop, RandomFlip, Normalize} \\
\midrule
\multicolumn{2}{l}{\textit{Optimizer}} \\
Optimizer Name & AdamW \\
Learning Rate (Initial) & $1 \times 10^{-3}$ \\
LR Scheduler & Cosine Annealing \\
Max Epochs & 10 \\
Warmup Epochs & 1 \\
\midrule
\multicolumn{2}{l}{\textit{DeAR Specific}} \\
Backbone & ViT-B/16 \\
Attribute Token Length ($K$) & 5 \\
Injection Layers & 9, 10, 11, 12 \\
Knowledge Anchoring $\beta$ & 0.9 \\
Regularization $\lambda_{\text{reg}}$ & 1.0 \\
Fusion Loss $\lambda_{\text{fusion}}$ & 0.7 \\
\bottomrule
\end{tabular}
\end{table}


\section{Functional Roles of Attention Heads}
\label{sec:appendix_head_roles}

To build the Role-Based Attention Mask for our DeAR framework, we systematically analyzed the top descriptive phrases for each attention head in the later layers (9-12) of the ViT-B/16 backbone. 

\paragraph{Offline Concept Entropy Analysis.} 
It is important to note that our framework does \textbf{not} rely on downstream task descriptions or labels for head discovery. Following TEXTSPAN, the Concept Entropy is computed via a one-time, offline analysis using a generic text corpus on the pre-trained backbone. The identified head roles are intrinsic to the model, kept frozen, and universally transferred to all downstream tasks without the need for re-discovery.

\paragraph{Q, K, V Masking Mechanism.} 
To strictly enforce functional separation, we implement the role isolation directly within the self-attention mechanism. Specifically, a mask matrix (containing $-\infty$ values) is applied to the Attention matrix before the Softmax operation. This definitively blocks the newly inserted Attribute Tokens from attending to (or being attended by) the original tokens associated with non-target heads, fully preserving the pre-trained backbone's innate generalization flow.

\paragraph{Conceptual Clusters and Role Assignment.} 
Our unsupervised analysis using TEXTSPAN and HDBSCAN identified \textbf{12 distinct conceptual clusters}: \textit{Location, Object, Color, Shape, Texture, Animal, Human, Text, Number, Style, Emotion, and Action}. We classify each head into one of four functional roles:
\begin{enumerate}[itemsep=0pt]
    \item \textbf{Core Attribute Heads:} Heads focusing on the five selected core concepts (\textit{Color, Shape, Texture, Object, Location}). These heads are assigned specific attribute tokens via Eq. 9.
    \item \textbf{Other Specialized Heads:} Heads focusing on specific distinct concepts that are not part of the core set (e.g., \textit{Animal, Human}). To prevent interference, they are isolated from the new attribute tokens (using Eq. 8).
    \item \textbf{Generalization Heads:} Heads focusing on abstract concepts or global composition (e.g., \textit{Style}). These are critical for zero-shot robustness and are also isolated.
    \item \textbf{Mixed Heads:} Heads with no single dominant focus, allowing unrestricted attention flow.
\end{enumerate}

Table~\ref{tab:head_role_details_long} details the classification for all heads in Layers 9-12.

\begin{longtable}{@{} p{0.1\textwidth} p{0.32\textwidth} p{0.55\textwidth} @{}}
\caption{\textbf{Detailed functional role classification for Attention Heads in Layers 9--12 of ViT-B/16.} We categorize each head into four roles: \textbf{Core Attribute}, \textbf{Other Specialized}, \textbf{Generalization}, and \textbf{Mixed}.} 
\label{tab:head_role_details_long} \\

\toprule
\textbf{Head} & \textbf{Role (Concept)} & \textbf{Representative Phrases} \\
\midrule
\endfirsthead

\multicolumn{3}{c}%
{{\bfseries \tablename\ \thetable{} -- continued from previous page}} \\
\toprule
\textbf{Head} & \textbf{Role (Concept)} & \textbf{Representative Phrases} \\
\midrule
\endhead

\midrule
\multicolumn{3}{r}{{Continued on next page}} \\
\bottomrule
\endfoot

\bottomrule
\endlastfoot

\multicolumn{3}{c}{\cellcolor{gray!10}\textit{\textbf{Layer 9}}} \\
(9, 0) & \textbf{Core Attribute} (Location) & \textit{Photo taken in Galápagos Islands, Photo taken in Okavango Delta, Bustling cityscape at night} \\
(9, 1) & \textbf{Core Attribute} (Location) & \textit{Picture taken in Bhutan, Photo taken in the Rub' al Khali, An image of Andorra} \\
(9, 2) & \textbf{Mixed} & \textit{A network of veins, Eyes, Photograph taken in a rustic barn, A wizard's hat} \\
(9, 3) & \textbf{Core Attribute} (Location) & \textit{Aerial view of a snowy landscape, Picture taken in the Swiss chocolate factories, Australian coral reef} \\
(9, 4) & \textbf{Other Specialized} (Emotion) & \textit{Sarcastic raised eyebrow, Intrigued facial expression, A photo of a young person} \\
(9, 5) & \textbf{Generalization} (Style) & \textit{Cinematic framing, Dramatic chiaroscuro photography, Retro-style poster design} \\
(9, 6) & \textbf{Mixed} & \textit{Futuristic biotechnology, Artwork featuring zebra stripe motifs, Surreal digital collage} \\
(9, 7) & \textbf{Core Attribute} (Texture) & \textit{Delicate ceramic patterns, Close-up of a textured synthetic rubber, Ethereal double exposure} \\
(9, 8) & \textbf{Other Specialized} (Text) & \textit{Artwork featuring Morse code typography, Bold graffiti, Film noir-inspired tones} \\
(9, 9) & \textbf{Core Attribute} (Location) & \textit{Bustling cityscape at night, Picture taken in a city park, Serene countryside sunrise} \\
(9, 10) & \textbf{Generalization} (Style) & \textit{Detailed illustration of a celestial body, Stark minimalism, Geometric tessellation} \\
(9, 11) & \textbf{Mixed} & \textit{An image of three subjects, A photo with the letter T, detailed reptile close-up} \\

\midrule
\multicolumn{3}{c}{\cellcolor{gray!10}\textit{\textbf{Layer 10}}} \\
(10, 0) & \textbf{Core Attribute} (Object) & \textit{Picture with a single domesticated animal, A trunk, A whisker, An irregular heptagon} \\
(10, 1) & \textbf{Core Attribute} (Color) & \textit{Earthy color tones, Pop art colors, Image with a pink color, Film noir-inspired tones} \\
(10, 2) & \textbf{Core Attribute} (Texture) & \textit{Photo of a furry animal, Close-up of a textured synthetic fabric, Marbleized design} \\
(10, 3) & \textbf{Core Attribute} (Location) & \textit{Sunlit meadow path, Crumbling and abandoned building, Vibrant city alley} \\
(10, 4) & \textbf{Other Specialized} (Human) & \textit{Image with a team of subjects, An image of two subjects, Crowded and bustling scene} \\
(10, 5) & \textbf{Core Attribute} (Object) & \textit{An image of a Fashion Designer, A photograph of a small object, Striking fashion stance} \\
(10, 6) & \textbf{Core Attribute} (Location) & \textit{Gritty urban street scene, Serene meadow landscape, Breathtaking canyons} \\
(10, 7) & \textbf{Generalization} (Style) & \textit{Playful reflections, Cubist still life painting, Stark minimalism, Timeless classic artwork} \\
(10, 8) & \textbf{Core Attribute} (Location) & \textit{Picture taken in a city park, Busy airport terminal, Urban alleyway} \\
(10, 9) & \textbf{Core Attribute} (Object) & \textit{An image of a dish, A photo of a young person, An image of fish, Colorful hot air balloons} \\
(10, 10) & \textbf{Core Attribute} (Shape) & \textit{Image with an owl, A regular octagon, Image with a futuristic time travel device} \\
(10, 11) & \textbf{Generalization} (Style) & \textit{Intricate pencil drawing, Unexpected symmetry, Futuristic transportation} \\

\midrule
\multicolumn{3}{c}{\cellcolor{gray!10}\textit{\textbf{Layer 11}}} \\
(11, 0) & \textbf{Core Attribute} (Object) & \textit{A bookmark, A laptop, A bowl, A skirt, A jacket, A phone, A regular octagon} \\
(11, 1) & \textbf{Core Attribute} (Color) & \textit{A platinum silver color, A gold color, A photo with the letter F, Image with a pink color} \\
(11, 2) & \textbf{Other Specialized} (Animal) & \textit{Playful animals, Flowers, An image with dogs, Image with a butterfly, An image with seagulls} \\
(11, 3) & \textbf{Core Attribute} (Location) & \textit{Picture snapped in the Alaskan mountains, Photo taken in Bangkok Thailand, Ocean} \\
(11, 4) & \textbf{Core Attribute} (Texture) & \textit{Photo of a furry animal, A bookmark, Close-up of a textured synthetic wood, A skirt} \\
(11, 5) & \textbf{Core Attribute} (Texture) & \textit{Artwork featuring Morse code typography, Delicate embroidery, Kaleidoscopic patterns} \\
(11, 6) & \textbf{Core Attribute} (Location) & \textit{Cultural exhibition, Photograph taken in a cozy cafe, Picture snapped in the Greek islands} \\
(11, 7) & \textbf{Other Specialized} (Action) & \textit{Joyful toddlers, Hands in an embrace, A paw, An image with pedestrians} \\
(11, 8) & \textbf{Generalization} (Style) & \textit{Anime style image, A high-resolution image, Artwork featuring abstract fractal patterns} \\
(11, 9) & \textbf{Core Attribute} (Object) & \textit{A necklace, A sock, A megaphone, A jacket, A fork, A belt, Precise clock mechanism} \\
(11, 10) & \textbf{Core Attribute} (Location) & \textit{Tranquil boating on a lake, Peaceful rural farmland, Secluded beach cove} \\
(11, 11) & \textbf{Core Attribute} (Location) & \textit{Bustling city nightlife, Secluded forest cabin, Energetic music festival crowd} \\

\midrule
\multicolumn{3}{c}{\cellcolor{gray!10}\textit{\textbf{Layer 12}}} \\
(12, 0) & \textbf{Core Attribute} (Location) & \textit{Picture taken in the Scotland countryside, Photo taken in Barcelona Spain, Photo taken in Tokyo} \\
(12, 1) & \textbf{Generalization} (Style) & \textit{Reflections, Central focal point, Motion freeze, Captivating city pulse, Dramatic skies} \\
(12, 2) & \textbf{Generalization} (Style) & \textit{Ephemeral glimmers, Whispering horizons, Symmetry disrupted, Unconventional beauty} \\
(12, 3) & \textbf{Mixed} & \textit{Picture with multiple wild animals, An image of a Chef de Cuisine, Picture with cars} \\
(12, 4) & \textbf{Other Specialized} (Human) & \textit{Playful siblings, Image with a five people, A photo of a woman, A group photo} \\
(12, 5) & \textbf{Core Attribute} (Shape) & \textit{Photo of a reptile, Image with a seagull, A scalene triangle, A snail, Herringbone pattern} \\
(12, 6) & \textbf{Core Attribute} (Location) & \textit{Photo taken in Namib Desert, Photo taken in the Alaskan mountains, Scottish moors} \\
(12, 7) & \textbf{Core Attribute} (Object) & \textit{A stick, A cup, A bonnet, A shoelace, A bottle, A belt, A jacket, A puddle} \\
(12, 8) & \textbf{Mixed} & \textit{A laptop, A rug, A shelf, A bag, Picture taken in an art gallery, Urban subway station} \\
(12, 9) & \textbf{Other Specialized} (Number) & \textit{An image of the number 10, An image of the number 7, The number fifteen} \\
(12, 10) & \textbf{Core Attribute} (Color) & \textit{An image with cold green tones, Image with a red color, A charcoal gray color} \\
(12, 11) & \textbf{Other Specialized} (Text) & \textit{A photo with the letter J, A photo with the letter K, A swirling eddy, A photo with the letter C} \\
\end{longtable}


\section{Complete Experiment Results}
\label{sec:appendix_results}

Our DeAR framework consistently outperforms previous approaches across all benchmark datasets. 

\subsection{Statistical Robustness and Variance}
To ensure the statistical reliability of our improvements, we evaluate DeAR alongside a strong baseline (MMRL++) across multiple random seeds ($\{1, 2, 3\}$). As presented in Table~\ref{tab:variance}, DeAR exhibits extremely low variance across independent runs and consistently outperforms the baseline on both the average of the 11 base-to-novel datasets and the challenging ImageNet-1K benchmark. This validates that the performance gains are robust and directly attributable to the proposed role-based masking strategy rather than initialization artifacts.

\begin{table}[h]
    \centering
    \caption{\textbf{Robustness Analysis.} Results are reported as Mean $\pm$ Standard Deviation across 3 independent seeds.}
    \vspace{0.2cm}
    \begin{tabular}{lcccccc}
        \toprule
        & \multicolumn{3}{c}{\textbf{Average (11 Datasets)}} & \multicolumn{3}{c}{\textbf{ImageNet-1K}} \\
        \cmidrule(lr){2-4} \cmidrule(lr){5-7}
        \textbf{Method} & \textbf{Base} & \textbf{Novel} & \textbf{HM} & \textbf{Base} & \textbf{Novel} & \textbf{HM} \\
        \midrule
        MMRL++ & 85.53 & 78.32 & 81.77 & \textbf{77.63} & 71.50 & 74.44 \\
        \rowcolor{gray!10} \textbf{DeAR (Ours)} & \textbf{85.62}{\tiny$\pm$0.21} & \textbf{79.77}{\tiny$\pm$0.11} & \textbf{82.71}{\tiny$\pm$0.18} & 77.52{\tiny$\pm$0.39} & \textbf{71.84}{\tiny$\pm$0.17} & \textbf{74.73}{\tiny$\pm$0.23} \\
        \bottomrule
    \end{tabular}
    \label{tab:variance}
\end{table}

\begin{figure}[h!]
\centering
\begin{subfigure}[t]{0.32\textwidth}
    \centering
    \includegraphics[width=\textwidth]{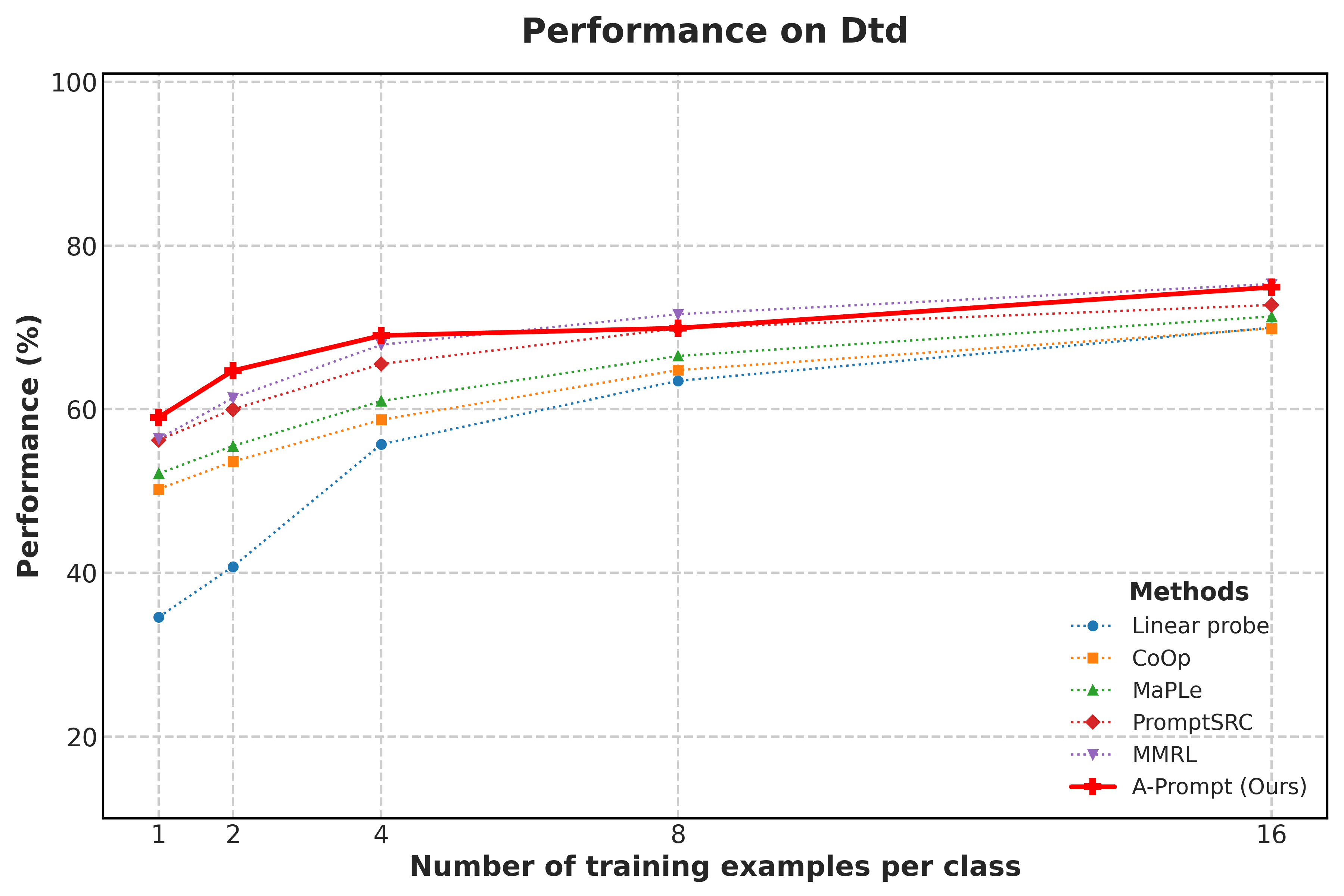}
    \caption{Average (11 datasets)}
\end{subfigure}
\hfill
\begin{subfigure}[t]{0.32\textwidth}
    \centering
    \includegraphics[width=\textwidth]{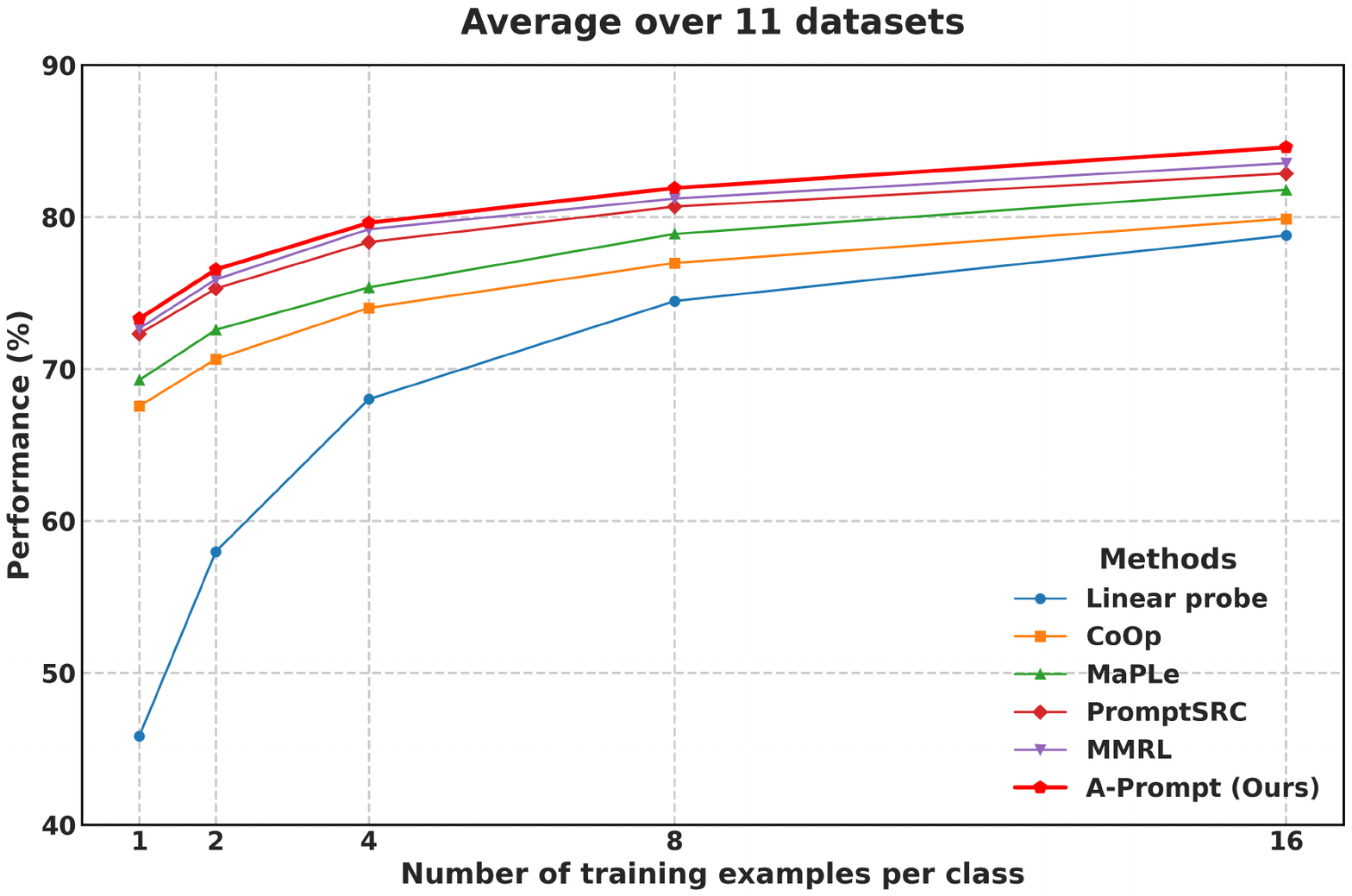}
    \caption{ImageNet}
\end{subfigure}
\hfill
\begin{subfigure}[t]{0.32\textwidth}
    \centering
    \includegraphics[width=\textwidth]{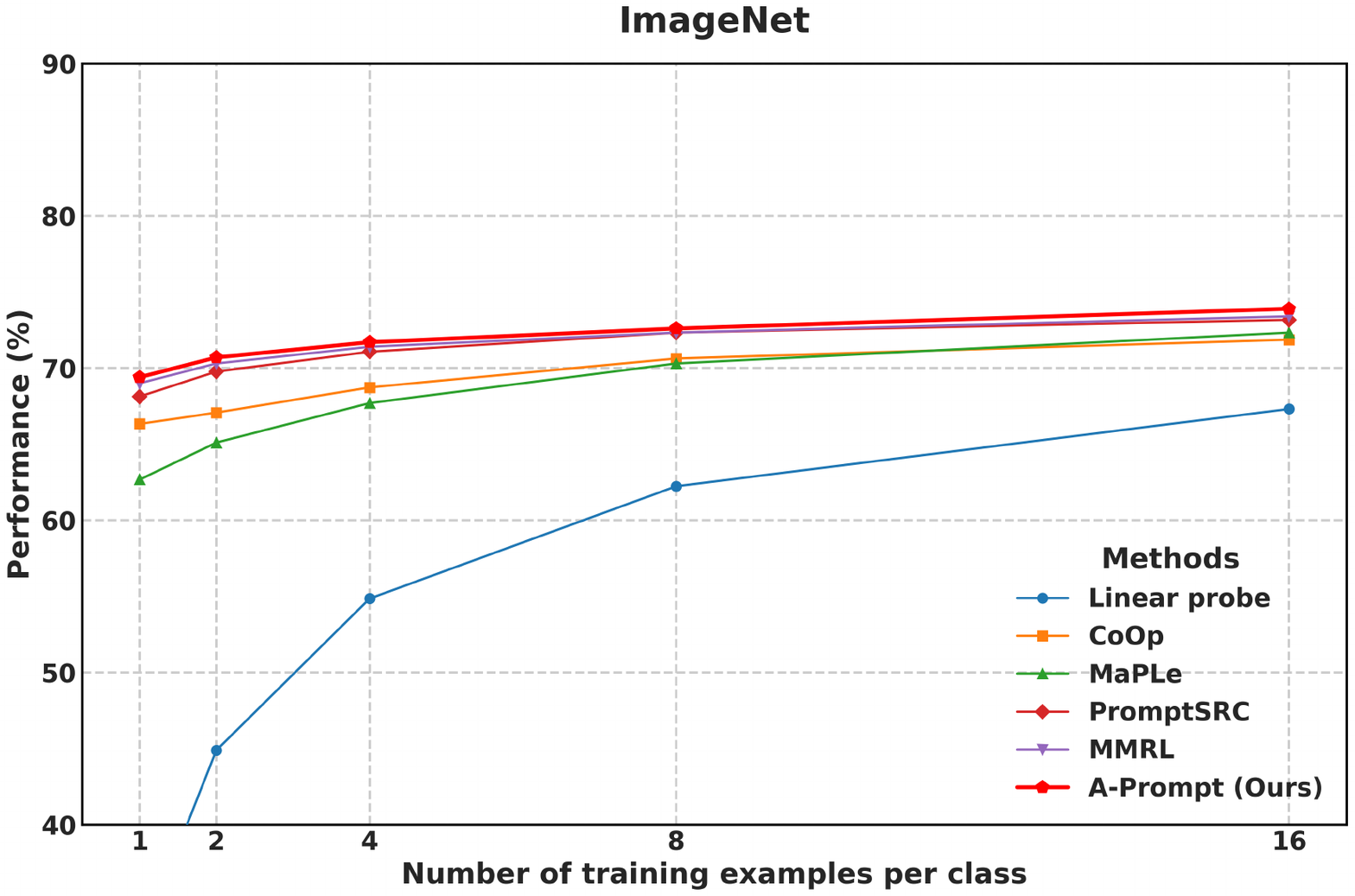}
    \caption{DTD}
\end{subfigure}

\vspace{0.2cm} 

\begin{subfigure}[t]{0.32\textwidth}
    \centering
    \includegraphics[width=\textwidth]{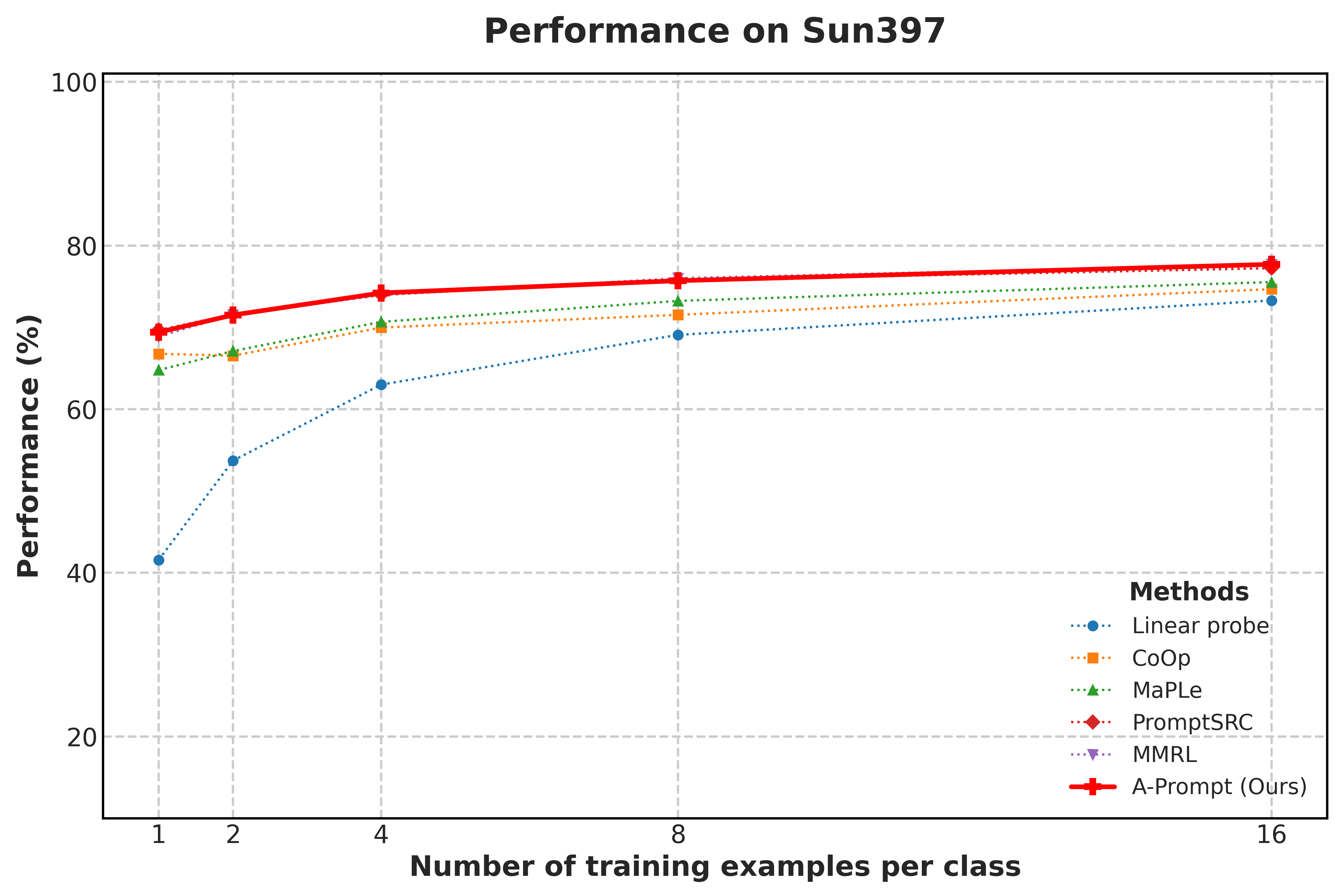}
    \caption{SUN397}
\end{subfigure}
\hfill
\begin{subfigure}[t]{0.32\textwidth}
    \centering
    \includegraphics[width=\textwidth]{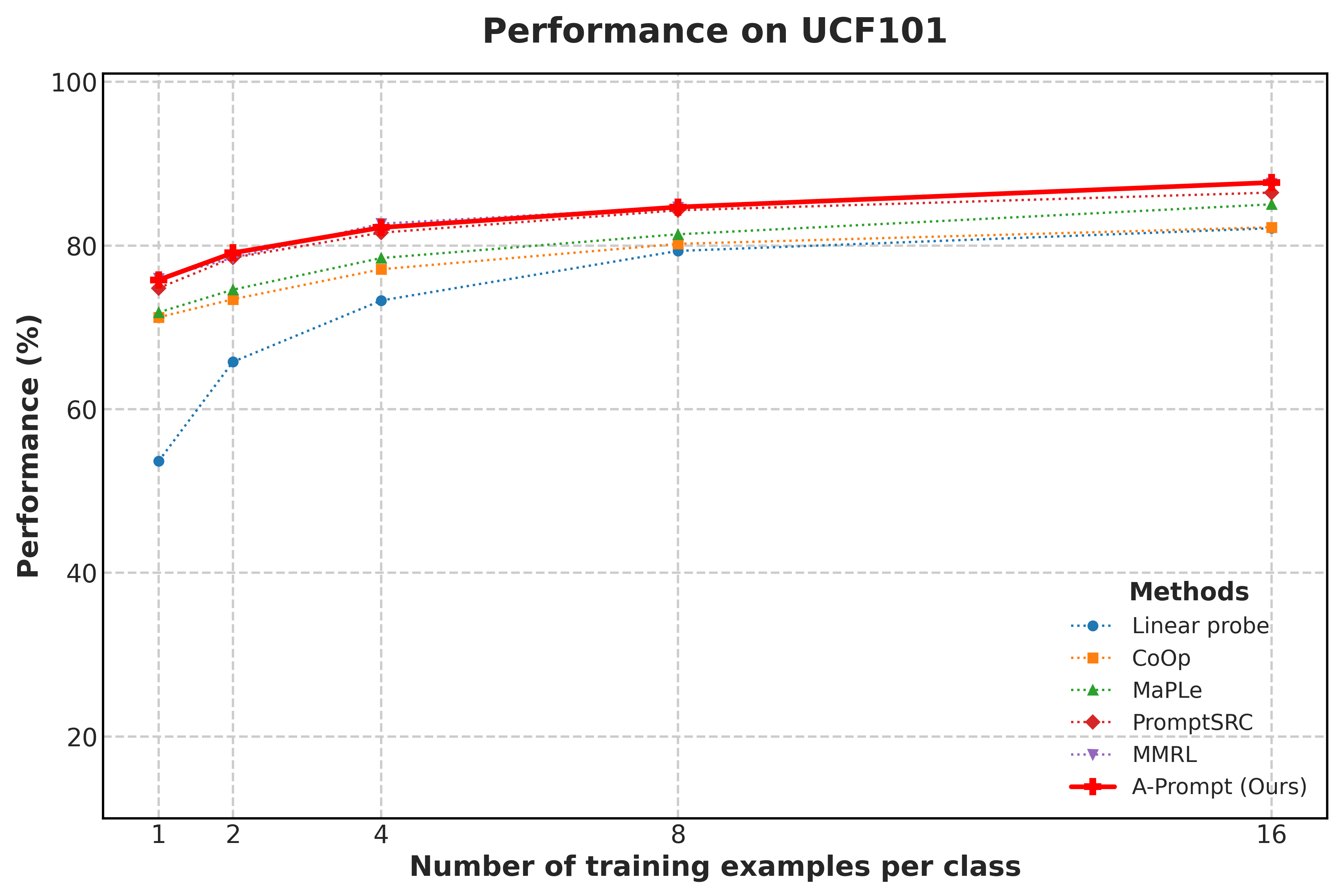}
    \caption{UCF101}
\end{subfigure}
\hfill
\begin{subfigure}[t]{0.32\textwidth}
    \centering
    \includegraphics[width=\textwidth]{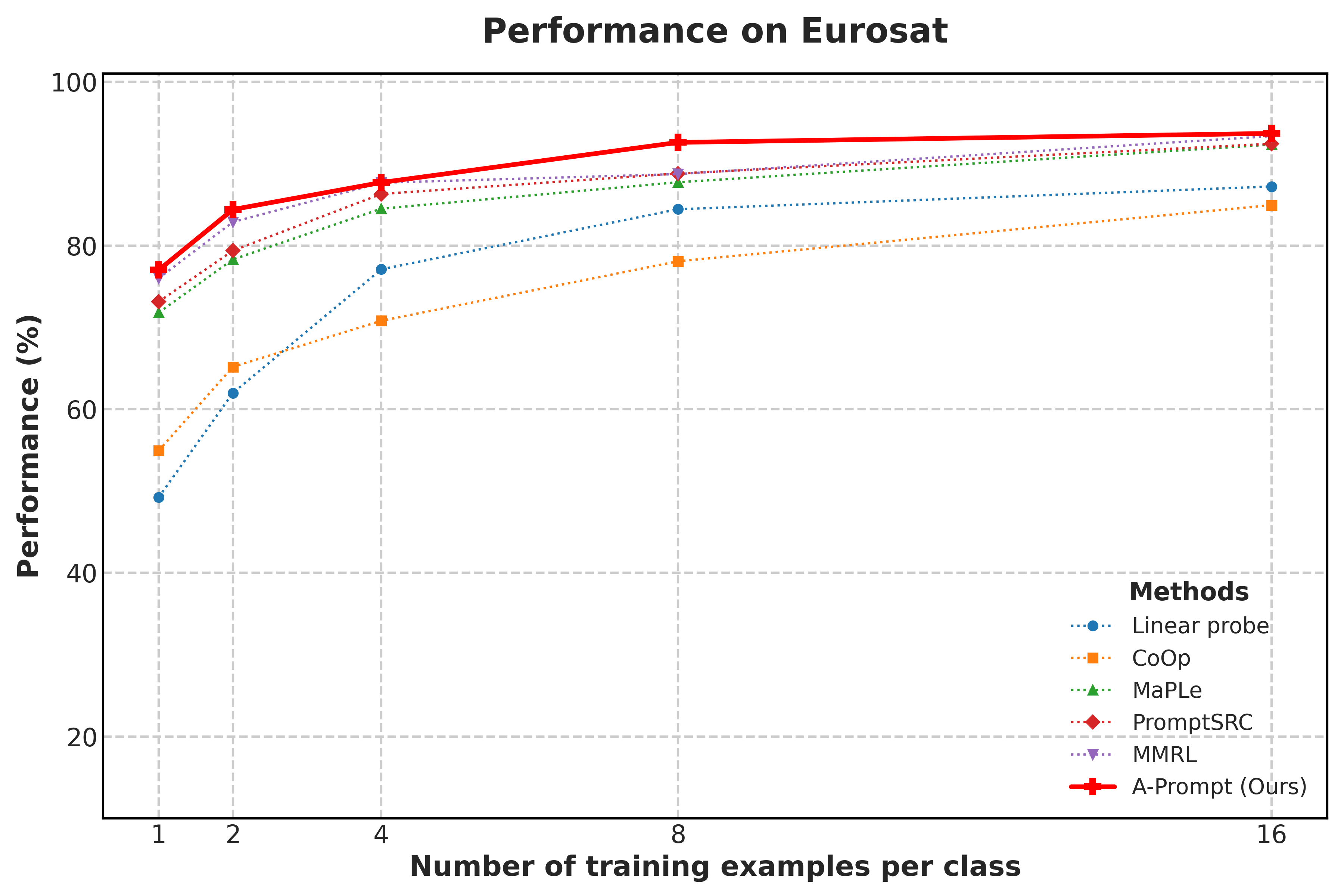}
    \caption{EuroSAT}
\end{subfigure}

\vspace{0.2cm}

\begin{subfigure}[t]{0.32\textwidth}
    \centering
    \includegraphics[width=\textwidth]{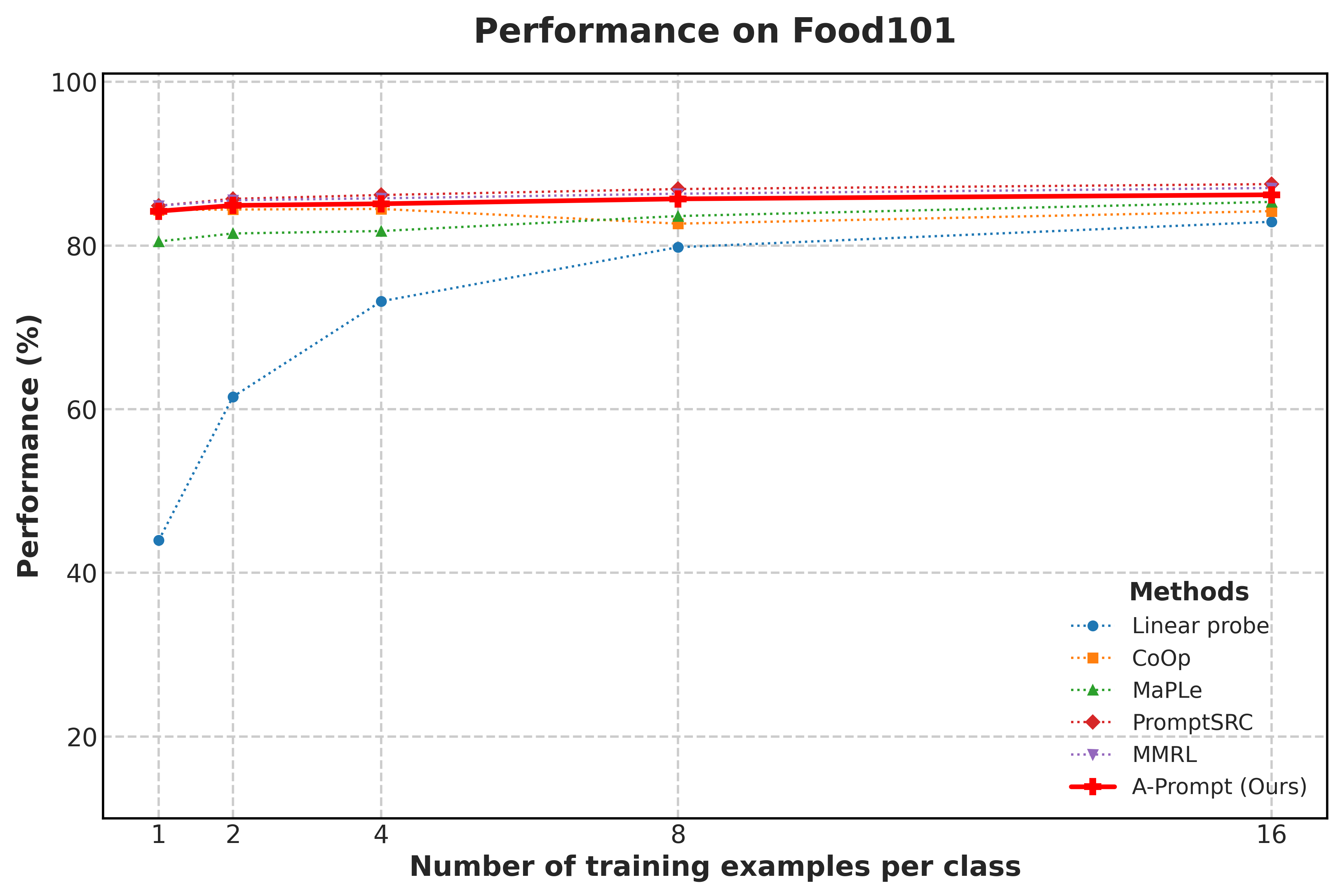}
    \caption{Food101}
\end{subfigure}
\hfill
\begin{subfigure}[t]{0.32\textwidth}
    \centering
    \includegraphics[width=\textwidth]{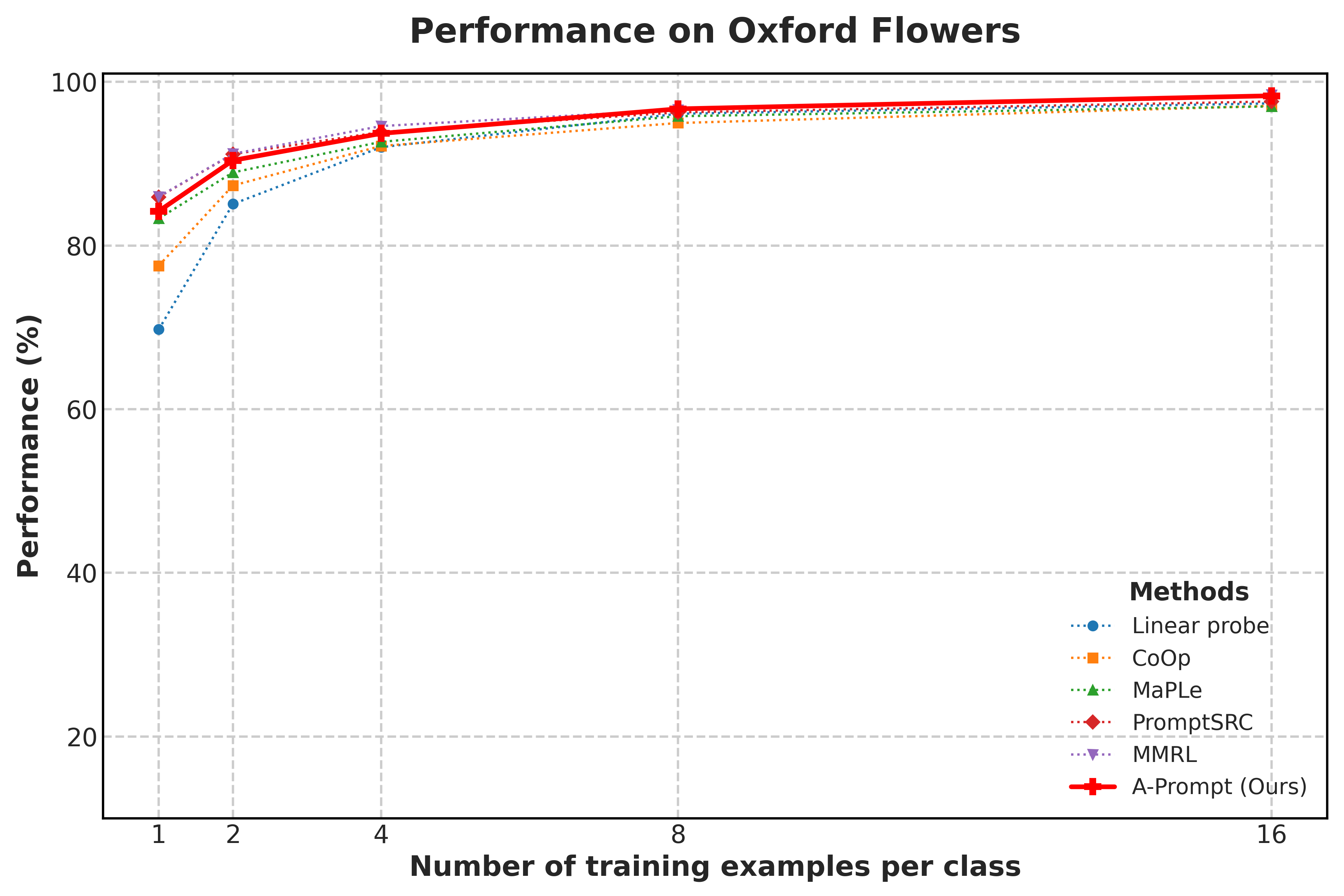}
    \caption{Oxford Flowers}
\end{subfigure}
\hfill
\begin{subfigure}[t]{0.32\textwidth}
    \centering
    \includegraphics[width=\textwidth]{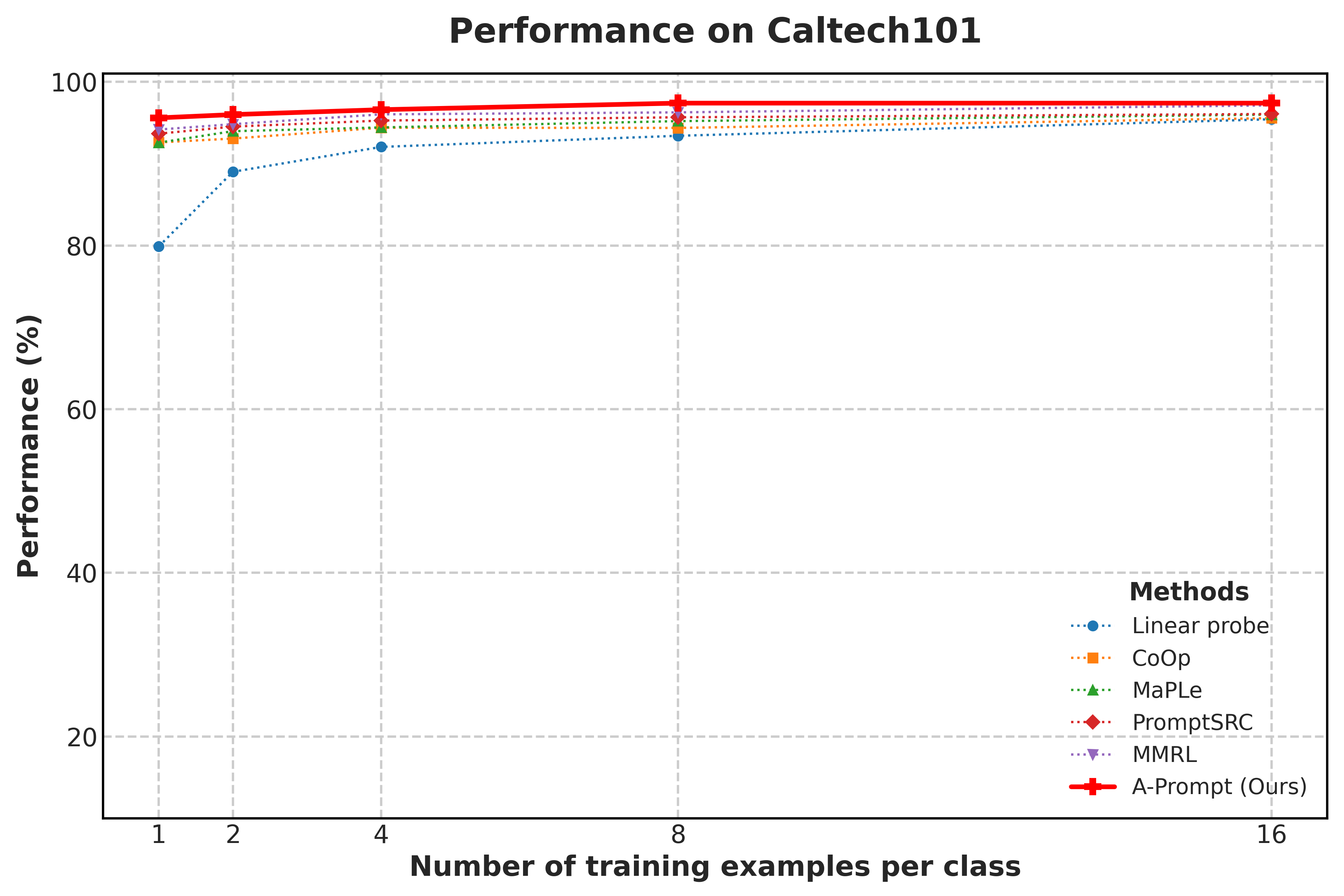}
    \caption{Caltech101}
\end{subfigure}

\vspace{0.2cm}

\begin{subfigure}[t]{0.32\textwidth}
    \centering
    \includegraphics[width=\textwidth]{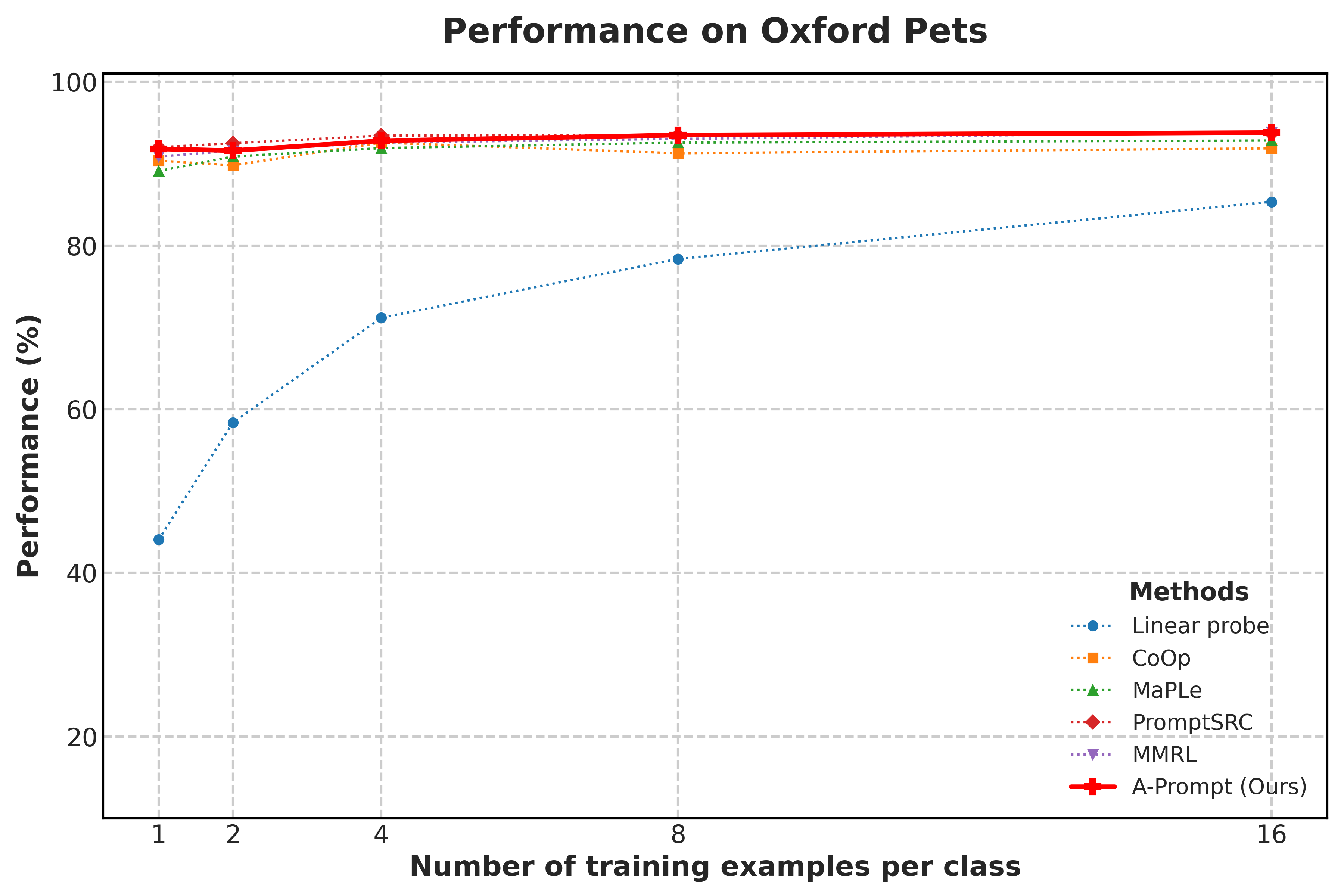}
    \caption{Oxford Pets}
\end{subfigure}
\hfill
\begin{subfigure}[t]{0.32\textwidth}
    \centering
    \includegraphics[width=\textwidth]{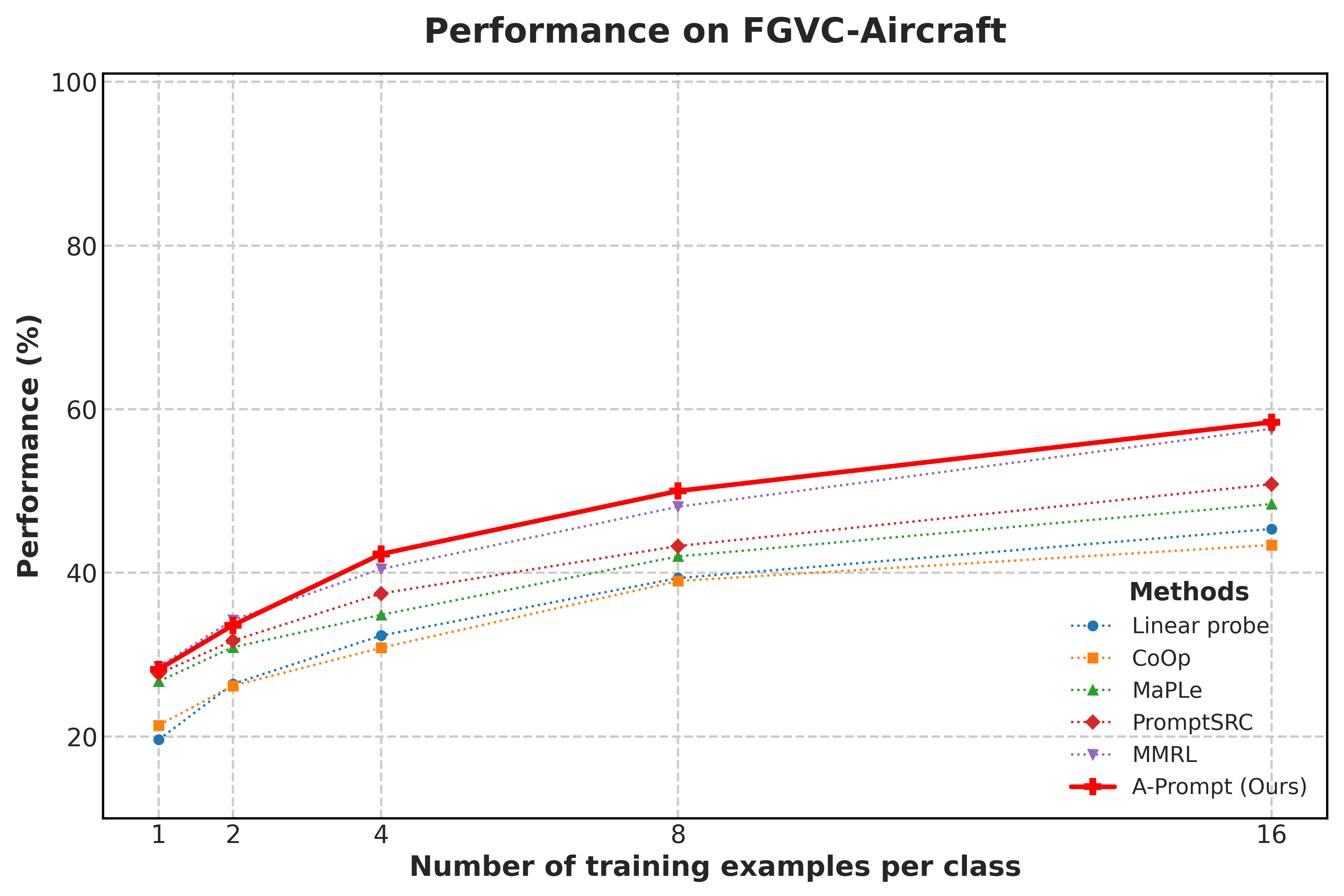}
    \caption{FGVC Aircraft}
\end{subfigure}
\hfill
\begin{subfigure}[t]{0.32\textwidth}
    \centering
    \includegraphics[width=\textwidth]{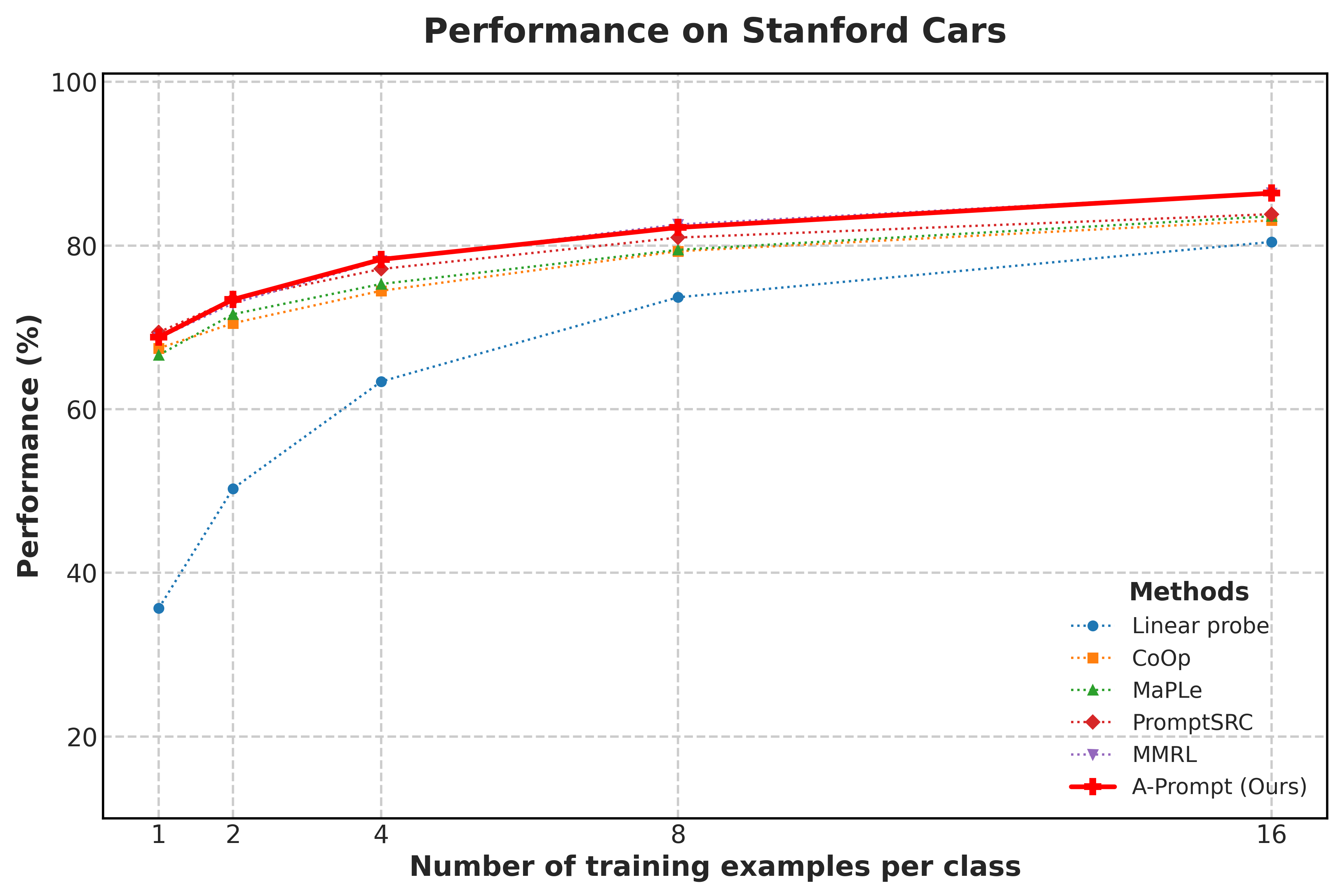}
    \caption{Stanford Cars}
\end{subfigure}

\caption{\textbf{Few-Shot Performance Comparison.} Comparison of DeAR with previous methods across 11 datasets. DeAR consistently achieves superior performance in few-shot settings.}
\label{fig:all_images}
\end{figure}

\section{Prompt Templates}
Table~\ref{tab:prompt_templates} summarizes the specific prompt templates used for each dataset in our experiments. These templates are manually designed to better match the characteristics of each dataset.

\begin{table}[h!]
\centering
\caption{Prompt templates used for each dataset.}
\label{tab:prompt_templates}
\begin{tabular}{ll}
\toprule
\textbf{Dataset} & \textbf{Prompt Template} \\
\midrule
OxfordPets         & a photo of a \{\}, a type of pet. \\
OxfordFlowers      & a photo of a \{\}, a type of flower. \\
FGVCAircraft       & a photo of a \{\}, a type of aircraft. \\
DescribableTextures& \{\} texture. \\
EuroSAT            & a centered satellite photo of \{\}. \\
StanfordCars       & a photo of a \{\}. \\
Food101            & a photo of \{\}, a type of food. \\
SUN397             & a photo of a \{\}. \\
Caltech101         & a photo of a \{\}. \\
UCF101             & a photo of a person doing \{\}. \\
ImageNet           & a photo of a \{\}. \\
ImageNetSketch     & a photo of a \{\}. \\
ImageNetV2         & a photo of a \{\}. \\
ImageNetA          & a photo of a \{\}. \\
ImageNetR          & a photo of a \{\}. \\
\bottomrule
\end{tabular}
\end{table}

\section{Additional Ablation Studies}

\paragraph{Task-Adaptive Fusion Logic and DTD Discrepancy.}
A key component of DeAR is its Task-Adaptive Fusion strategy. The purpose of $\mathcal{L}_{\text{fusion}}$ is to encourage the model to prioritize the generalization feature ($\mathbf{f}_{\text{cls}}$) as the primary decision-maker, utilizing the specialized attributes ($\mathbf{f}_{\text{attr}}$) solely for fine-grained refinement. Without this loss constraint, the model risks over-relying on learned attributes, which degrades zero-shot capability. 

It is worth noting the behavioral difference on the Describable Textures Dataset (DTD). In the base-to-novel setting, predictions utilize the full Task-Adaptive Fusion ($\mathbf{f}_{\text{attr}} + \mathbf{f}_{\text{cls}}$), which heavily benefits texture-centric datasets like DTD. Conversely, in the cross-dataset evaluation protocol, predictions rely solely on the decoupled general-purpose feature ($\mathbf{f}_{\text{cls}}$) for strict fairness across methods. This discrepancy actually validates that our attribute tokens successfully capture critical fine-grained features that $\mathbf{f}_{\text{cls}}$ alone might miss.

As shown in Table~\ref{tab:ablation_inference}, our proposed Task-Adaptive Fusion significantly outperforms a standard decoupled strategy (+1.23\% HM).

\begin{table}[h!]
\centering
\caption{Ablation study on the inference strategy. Average accuracy (\%) on the base-to-novel benchmark.}
\label{tab:ablation_inference}
\begin{tabular}{lccc}
\toprule
\textbf{Inference Strategy} & \textbf{Base} & \textbf{Novel} & \textbf{HM} \\
\midrule
Decoupled Inference & \textbf{85.94} & 78.50 & 82.05 \\
\textbf{Task-Adaptive Fusion (Ours)} & \textbf{85.94} & \textbf{79.73} & \textbf{82.72} \\
\bottomrule
\end{tabular}
\end{table}

\paragraph{Threshold Sensitivity and Role Ablation.}
The Role-Based Mask uses Concept Entropy quantiles to separate specialized heads (bottom 20\% entropy) from generic heads (top 20\% entropy). To demonstrate the stability of this heuristic, we conduct a sensitivity analysis on ImageNet-1K (Table~\ref{tab:ablation_combined}, Left). DeAR maintains a stable performance plateau across a broad threshold range (Top 10\% to 30\%), indicating it does not require fragile hyperparameter tuning.

Furthermore, accurately identifying these roles is paramount. As shown in Table~\ref{tab:ablation_combined} (Right), randomly shuffling the assigned roles results in a severe performance drop (from 74.73\% to 70.85\%), conclusively proving the necessity of the proposed entropy-based role decomposition. 

\begin{table}[h]
    \centering
    \caption{\textbf{Sensitivity Analysis and Ablation Studies on ImageNet-1K.}}
    \vspace{0.2cm}
    \begin{tabular}{lc|lc}
        \toprule
        \multicolumn{2}{c|}{\textbf{Sensitivity (Entropy Threshold \%)}} & \multicolumn{2}{c}{\textbf{Role \& Attribute ($K$) Ablation}} \\
        \textbf{Threshold Q.} & \textbf{IN-1K HM ($\pm$Std)} & \textbf{Strategy} & \textbf{IN-1K HM ($\pm$Std)} \\
        \midrule
        Top 10\%  & 74.35 $\pm$ 0.21 & Random Role Shuffle & 70.85 $\pm$ 0.25 \\
        \rowcolor{gray!10} \textbf{Top 20\%} & \textbf{74.73 $\pm$ 0.23} & \textbf{DeAR ($K=5$)} & \textbf{74.73 $\pm$ 0.23} \\
        Top 30\%  & 74.28 $\pm$ 0.20 & Concept Cluster $K=3$ & 73.85 $\pm$ 0.29 \\ 
        Top 40\%  & 71.10 $\pm$ 0.24 & Concept Cluster $K=7$ & 74.01 $\pm$ 0.33 \\ 
        \bottomrule
    \end{tabular}
    \label{tab:ablation_combined}
\end{table}

\paragraph{Sensitivity to "Core Attribute" Selection.}
Our default design selects $K=5$ core attributes. Table~\ref{tab:ablation_combined} (Right) also explores adjusting the granularity of these concept clusters. While $K=3$ or $K=7$ slightly underperform the optimal $K=5$, they still substantially exceed baseline performances, demonstrating the framework's robustness to attribute granularity.

\begin{figure}[h!]
    \centering
    \includegraphics[width=0.85\textwidth]{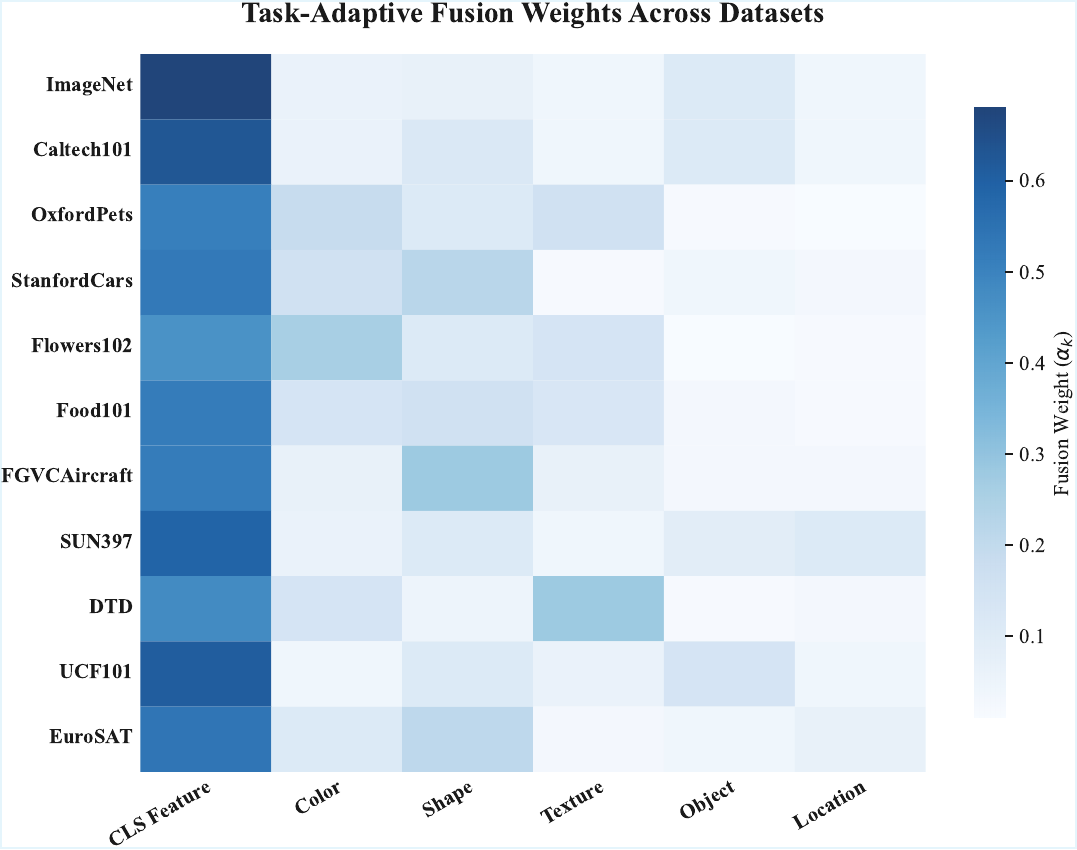} 
    \caption{\textbf{Visualization of Task-Adaptive Fusion Weights Across Datasets.} The heatmap displays the magnitude of the learned fusion weights ($\alpha_k$) for the global [CLS] feature and the five specialized attribute tokens. Darker blue indicates higher importance.}
    \label{fig:fusion_heatmap}
\end{figure}

\paragraph{Visualization of Learned Task Priors.} 
To further understand how DeAR adapts to different domains, we visualize the learned fusion weights $\alpha_k$ (averaged over heads of the same attribute type) across all 11 datasets in Figure~\ref{fig:fusion_heatmap}. 

The heatmap reveals highly interpretable patterns:
\begin{itemize}[itemsep=0pt]
    \item \textbf{DTD (Texture Dataset):} Shows a significantly higher weight for the \textbf{Texture} token.
    \item \textbf{OxfordPets \& Flowers102:} Exhibit balanced but elevated attention to \textbf{Color} and \textbf{Shape}.
    \item \textbf{FGVCAircraft \& StanfordCars:} Show reduced reliance on Color but maintain attention on \textbf{Shape}.
    \item \textbf{ImageNet:} Maintains a strong reliance on the global \textbf{[CLS] Feature}.
\end{itemize}

\section{Code}
Code is available at our \href{https://github.com/wellsssssss/DeAR}{GitHub repository}.